%% file: main.tex
\newif\ifRAL
\RALtrue 	% RAL PAPER
\ifRAL
  \documentclass[letterpaper, 10 pt, journal, twoside]{IEEEtran}
\else
  \documentclass[letterpaper, 10 pt, conference]{ieeeconf}
\fi

\input{./macros/boldfacemath}

\usepackage[utf8]{inputenc} % usually not needed (loaded by default)
\usepackage[T1]{fontenc}
\usepackage{graphicx}
\usepackage{array}
\usepackage{url}
\usepackage{amsmath}
\usepackage{amssymb}
\usepackage{amsthm}
\usepackage{amsfonts}
\usepackage{mathtools}
\usepackage{bm}
\usepackage{dsfont}
\usepackage[usenames,dvipsnames,table]{xcolor}
\usepackage{hyperref} %pdf with links and toc on the left
\hypersetup{
    colorlinks,
    citecolor=black,
    filecolor=black,
    linkcolor=black,
    urlcolor=black,
    pdfauthor={},
    pdfsubject={},
    pdftitle={}
}

\usepackage{pifont}% http://ctan.org/pkg/pifont

\usepackage{times}

\usepackage[normalem]{ulem} % pkg used to produce the \underline and \overline commands 

\usepackage[ruled,vlined,linesnumbered]{algorithm2e}
\usepackage[noend]{algorithmic}

\usepackage{subfiles}
\makeatletter
\def  \input@path{{./figures/tikz/}}
\makeatother

\usepackage[per-mode=symbol,binary-units]{siunitx}
\usepackage{cite} %this pkg has to be de-activated when using SAGE template

\graphicspath{{./figures/}{./figures/Plots/}}

\usepackage{verbatim}
\usepackage{latexsym}
\usepackage{booktabs}
\usepackage{multirow}	% multi-row
\usepackage{epsfig,epstopdf}
\usepackage[inline]{enumitem} % to use enum in line with custom item symbols

\usepackage{xparse}
\usepackage[textsize=scriptsize]{todonotes}

\usepackage[nomain,acronyms]{glossaries}

\usepackage{marginnote}

\def\baselinestretch{1.0}

\usepackage[ruled,vlined,linesnumbered]{algorithm2e}	%DB: norelsize option removed to avoid option clash
\SetCommentSty{textit}
\SetAlgoInsideSkip{smallskip}
\input{./macros/glossaries.tex}

\newcolumntype{C}{>{$}c<{$}} % math-mode version of "c" column type
	
\newcommand{\ie}{\textit{i}.\textit{e}.,\@\xspace}
\newcommand{\eg}{\textit{e}.\textit{g}.,\@\xspace}

\newcommand{\ith}{\textit{i}-\textit{th} } %the space character is left on purpose 
\newcommand{\figref}[1]{Fig.~\ref{#1}}
\newcommand{\tabref}[1]{Tab.~\ref{#1}}
\newcommand{\secref}[1]{Sec.~\ref{#1}}

\newcommand\figWidth{1}

\input{symbol_definitons}

\def\figFASTHexProto{\includegraphics[width=0.90\columnwidth]{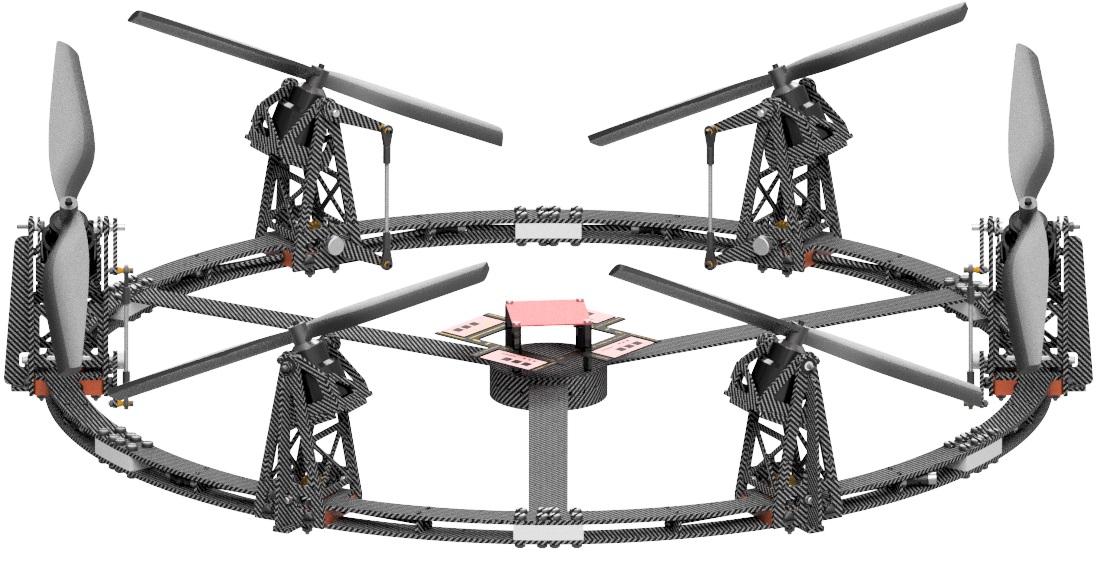}}
\def\figFASTHex{\includegraphics[width=0.90\columnwidth]{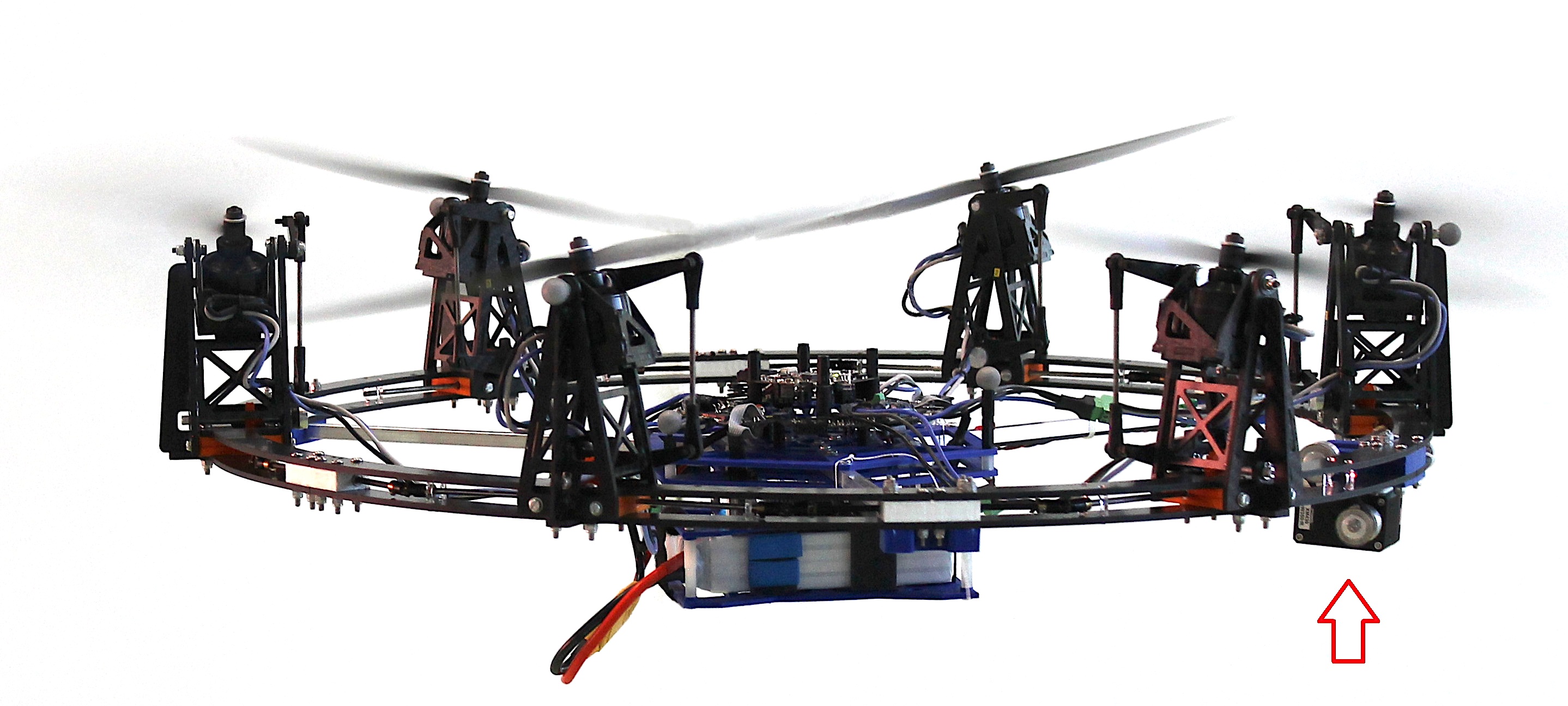}}
\def\figFASTHexRing{\includegraphics[width=0.90\columnwidth]{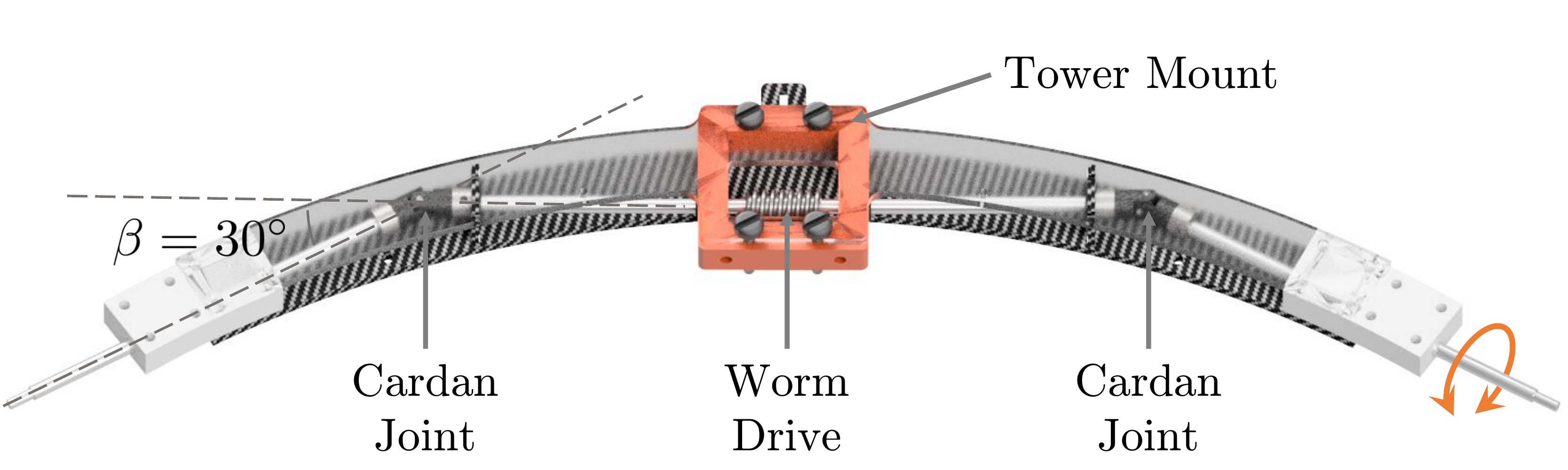}}
\def\figFASTHexTower{\includegraphics[width=0.80\columnwidth]{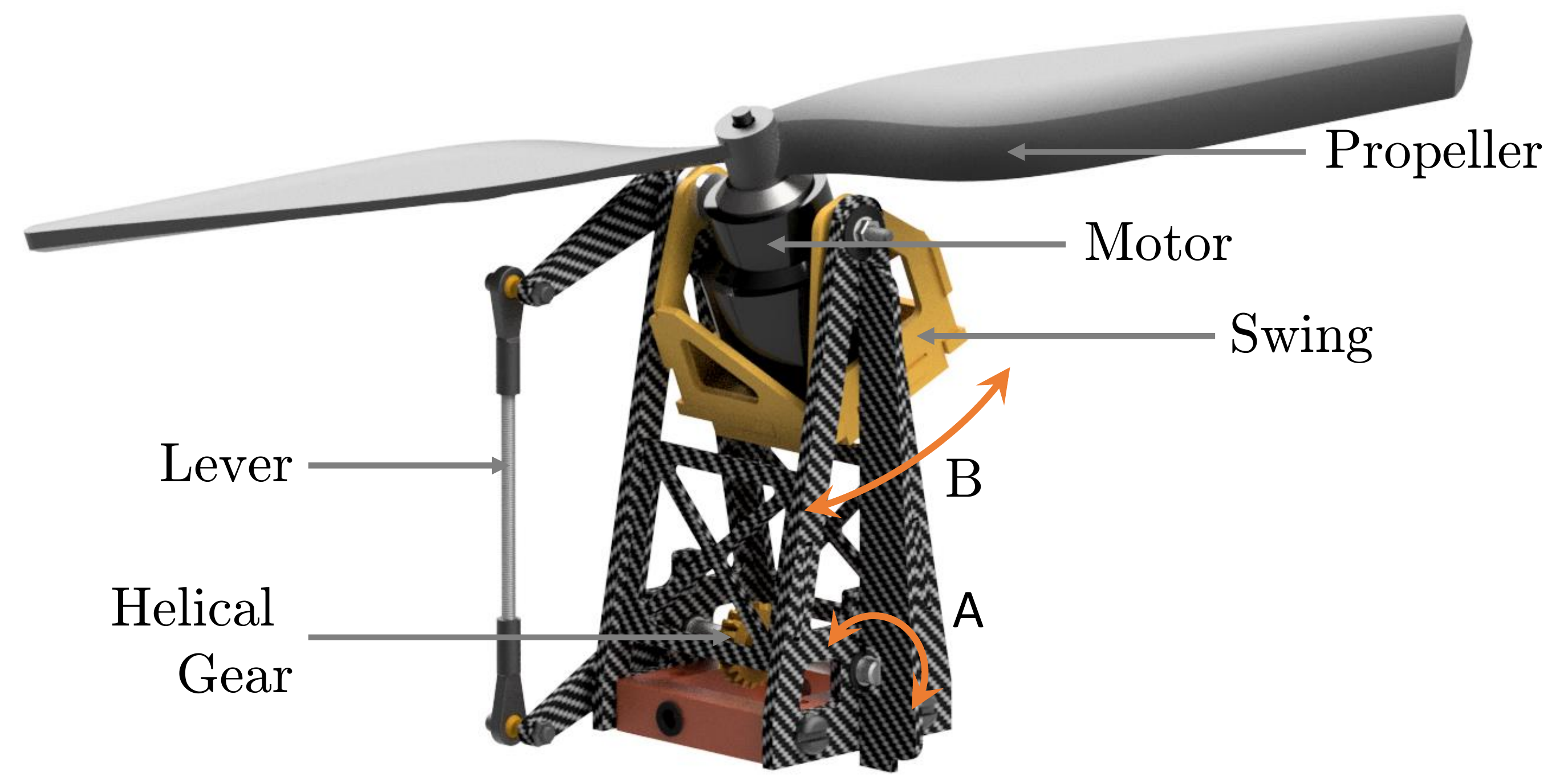}}
\def\figCardJointAngle{\includegraphics[width=0.9\columnwidth]{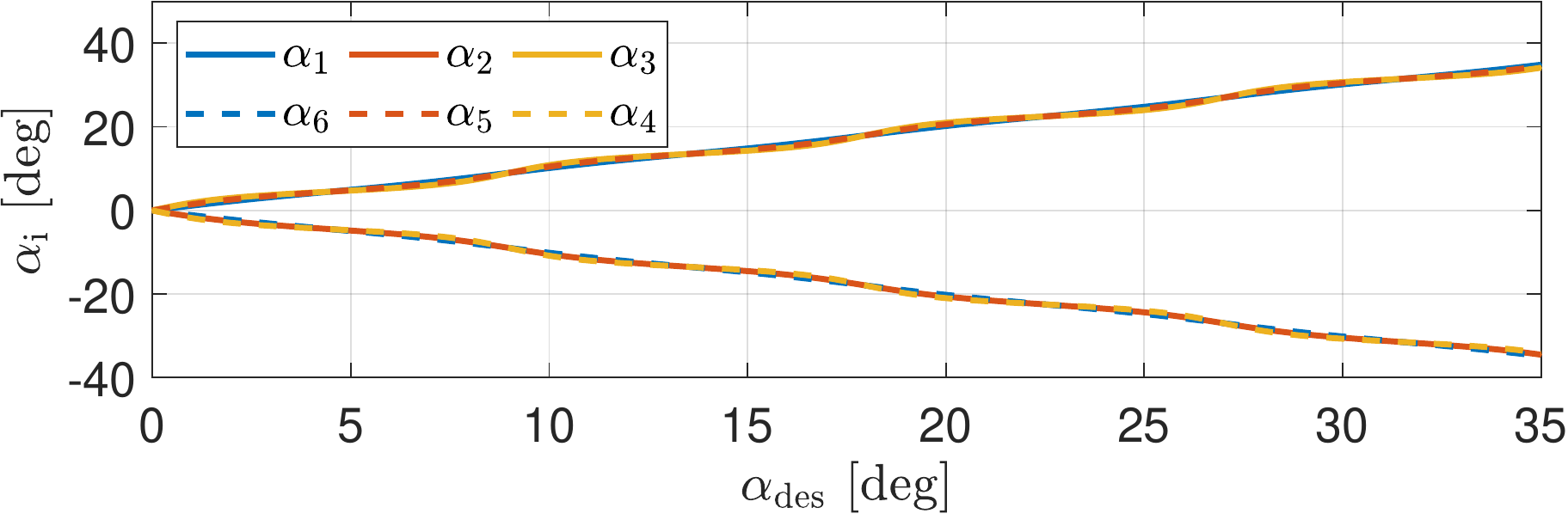}}

\def\figSchematic{\includegraphics[width=1.0\columnwidth]{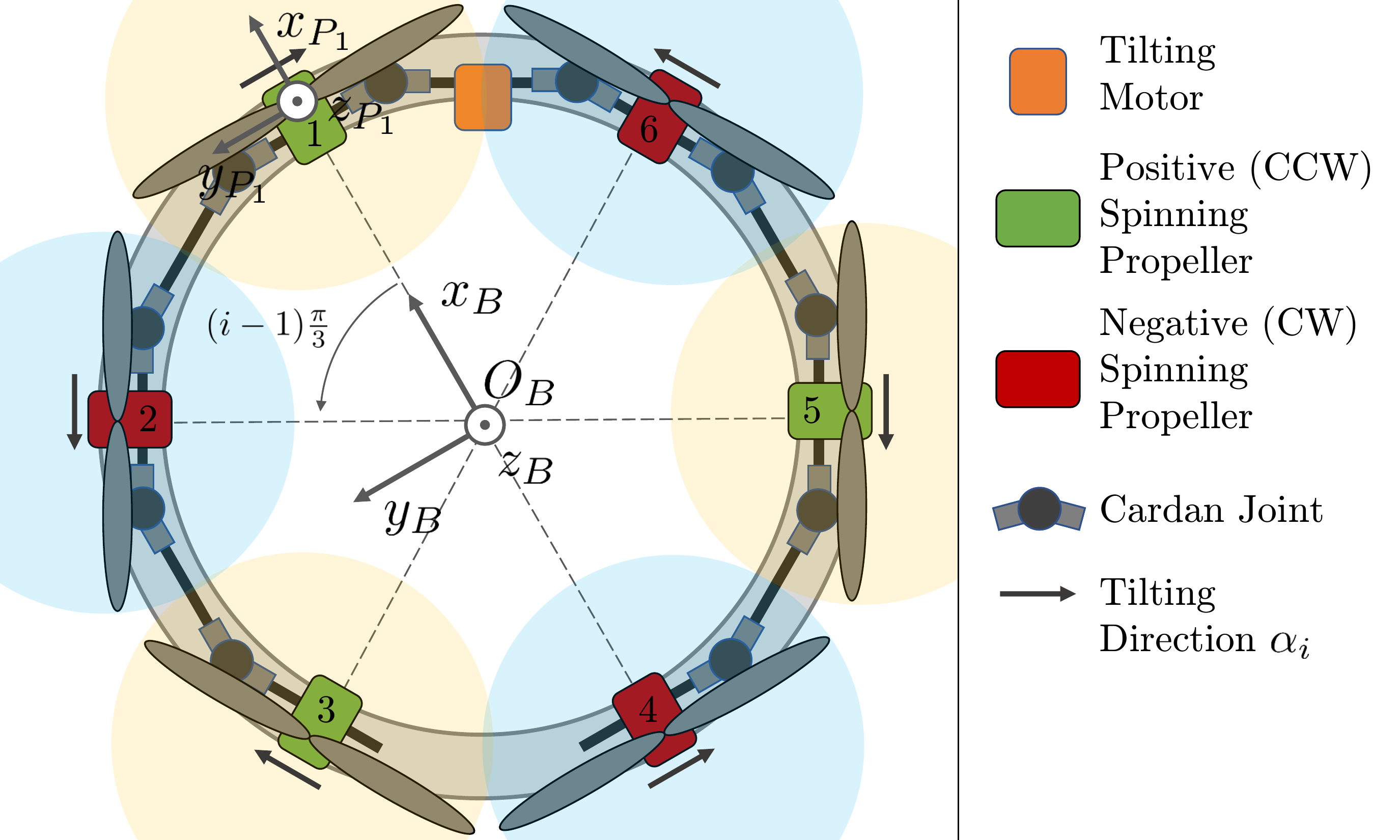}}

\def\figEfficiency{\includegraphics[width=1.0\columnwidth]{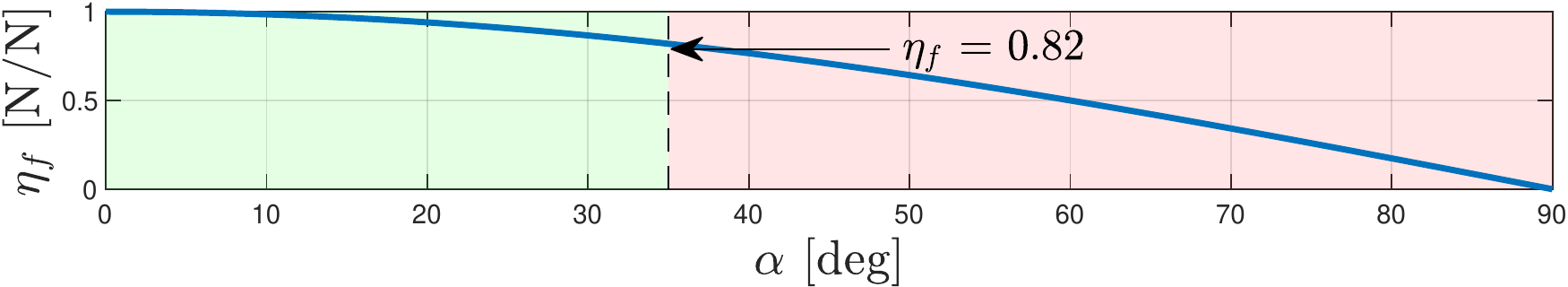}}
\def\figForceSpace{\includegraphics[width=1.0\columnwidth]{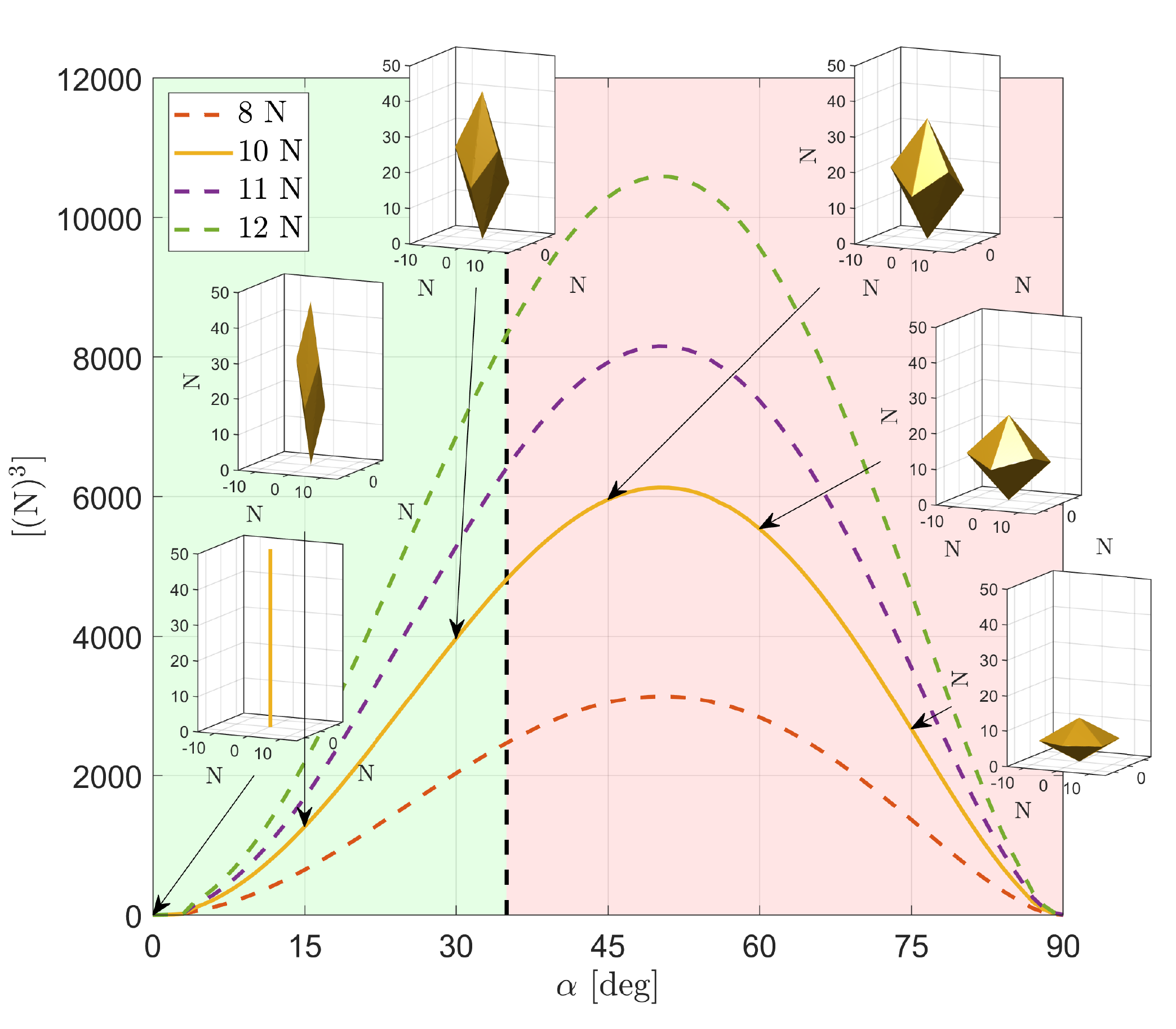}}
\def\figTorqueSpace{\includegraphics[width=1.0\columnwidth]{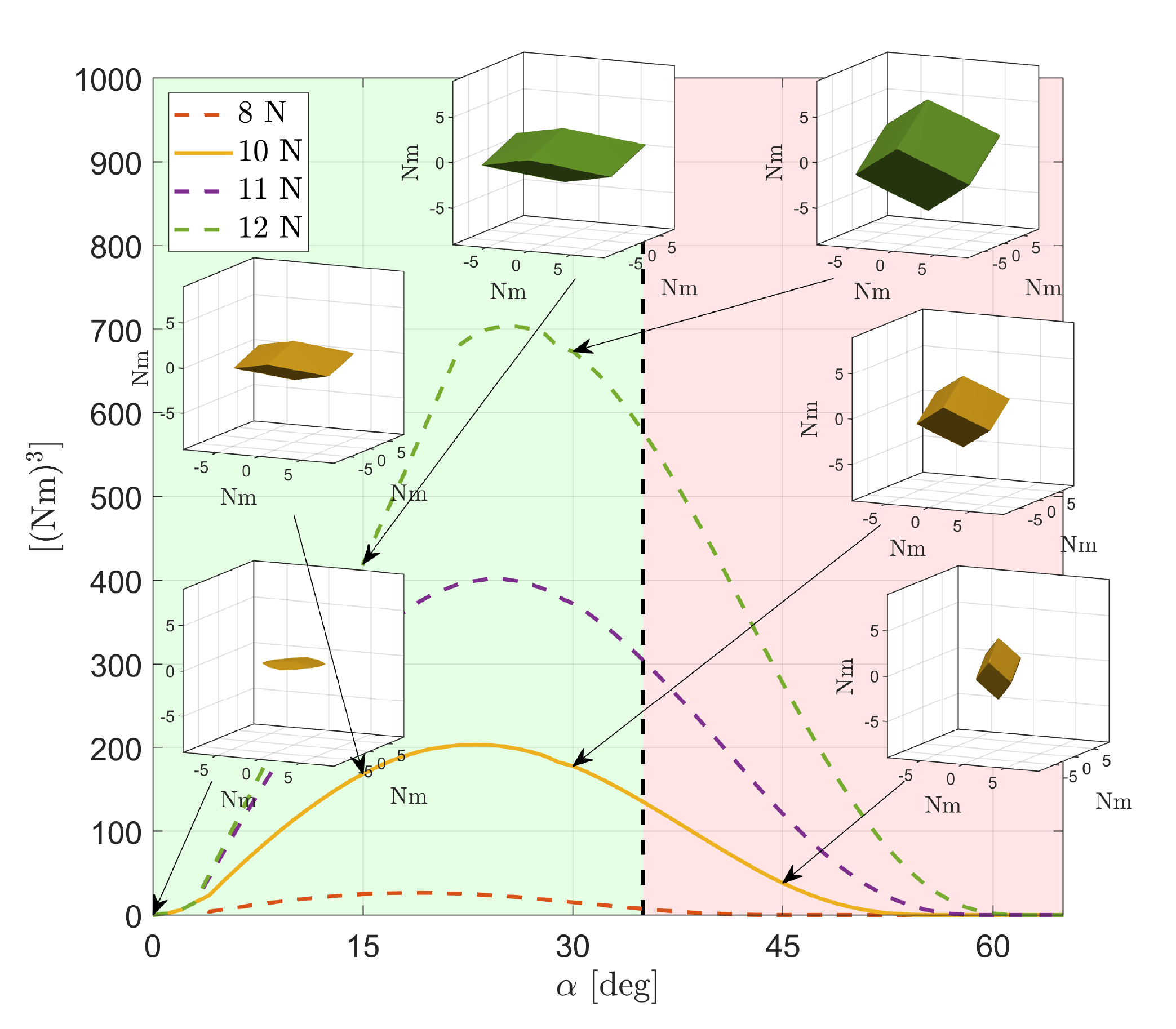}}

\def\figExp2_Image{\includegraphics[width=1\columnwidth]{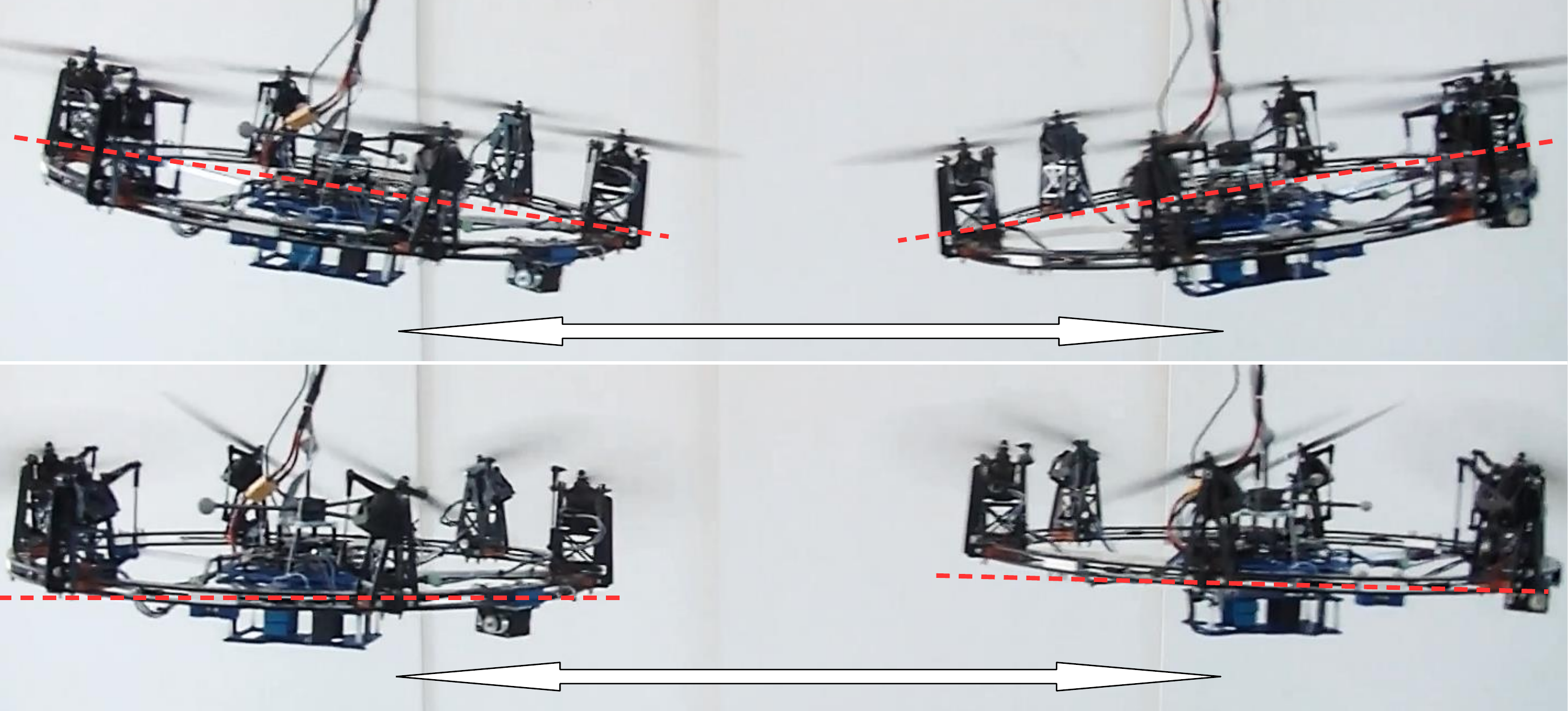}}

\title{\LARGE \bf FAST-Hex -- A Morphing Hexarotor: Design, Mechanical Implementation, Control and Experimental Validation}

\renewenvironment{IEEEbiography}[1]
  {\IEEEbiographynophoto{#1}}
  {\endIEEEbiographynophoto}

\author{Markus Ryll$^1$, Davide Bicego$^{4,2}$, Mattia Giurato$^3$, Marco Lovera$^3$ and Antonio Franchi$^{4,2}$
\thanks{\scriptsize{$^{1}$Computer Science and Artificial Intelligence Laboratory, Massachusetts Institute of Technology, Cambridge, USA, \href{mailto:ryll@mit.edu}{ryll@mit.edu}}}
\thanks{\scriptsize{$^2$LAAS-CNRS, Universit\'e de Toulouse, CNRS, Toulouse, France}}
\thanks{\scriptsize{$^3$Dipartimento di Scienze e Tecnologie Aerospaziali, Politecnico di Milano, Milano, Italy, \href{mailto:mattia.giurato@polimi.it}{mattia.giurato@polimi.it}}}
\thanks{\scriptsize{$^4$Robotics and Mechatronics lab, Faculty of Electrical Engineering, Mathematics \& Computer Science, University of Twente, Enschede, The Netherlands,  \href{a.franchi@utwente.nl}{\{d.bicego, a.franchi\}@utwente.nl}}}
\thanks{\scriptsize{This work has been partially funded by the European Union's Horizon 2020 research and innovation programme under grant agreement  ID: 871479  AERIAL-CORE.}}
}

\makeatletter
\def\ps@titlepagestyle{
	\def\@oddfoot{\textcolor{green}{\sf\footnotesize Preprint version, final version under review \hfill}}
	\def\@evenfoot{\textcolor{green}{\sf\footnotesize Preprint version, final version under review \hfill}}
	\def\@oddhead{\textcolor{red}{\sf\footnotesize Preprint version, final version under review \hfill}}
	\def\@evenhead{\textcolor{yellow}{\sf\footnotesize Preprint version, final version under review \hfill}}
}%
\makeatother
\makeatletter
\def\ps@headings{
	\def\@oddfoot{\textcolor{black}{\sf\footnotesize Preprint version, final version under review \hfill}}
	\def\@evenfoot{\textcolor{black}{\sf\footnotesize Preprint version, final version under review \hfill}}
}%
\makeatother
\begin{document}

\maketitle
\thispagestyle{headings}

%%%%%%%%%%%%%%%%%%%%%%%%%%%%%%%%%%%%%%%%%%%%%%%%%%%%%%%%%%%%%%%%%%%%%%
\begin{abstract}
We present FAST-Hex, a micro aerial hexarotor platform that allows to seamlessly transit from an \emph{under-actuated} to a \emph{fully-actuated} configuration with only one additional control input, a motor that synchronously tilts all propellers. The FAST-Hex adapts its configuration between the more efficient but under-actuated, collinear multi-rotors and the less efficient, but full-pose-tracking, which is attained by non-collinear multi-rotors. On the basis of prior work on minimal input configurable micro aerial vehicle we mainly stress three aspects: mechanical design, motion control and experimental validation. Specifically, we present the lightweight mechanical structure of the FAST-Hex that allows it  to only use one additional input to achieve configurability and full actuation in a vast state space. The motion controller receives as input any reference pose in $\mathbb{R}^3\times \mathrm{SO}(3)$ (3D position + 3D orientation). Full pose tracking is achieved if the reference pose is feasible with respect to actuator constraints. In case of unfeasibility a new feasible desired trajectory is generated online giving priority  to the position tracking over the orientation tracking. Finally we present a large set of experimental results shading light on all aspects of the control and pose tracking of FAST-Hex.
\end{abstract}
\vspace{-3mm}
%%%%%%%%%%%%%%%%%%%%%%%%%%%%%%%%%%%%%%%%%%%%%%%%%%%%%%%%%%%%%%%%%%%%%%

%%%%%%%%%%%%%%%%%%%%%%%%%%%%%%%%%%%%%%%%%%%%%%%%%%%%%%%%%%%%%%%%%%%%%%
\section{INTRODUCTION}\label{sec:intro}
%%%%%%%%%%%%%%%%%%%%%%%%%%%%%%%%%%%%%%%%%%%%%%%%%%%%%%%%%%%%%%%%%%%%%%
Unmanned aerial vehicles (UAVs) are used in a wide spectrum of applications like environmental and infrastructural monitoring and aerial photography, search and rescue operations and aerial physical interaction, including transportation, sensing by contact and assembly tasks, just to name a few. 
These very different applications resulted in a broad potpourri of differently shaped UAVs. 
For high altitude, long duration surveillance applications a fixed-wing UAV is the optimal candidate. 
For applications in confined and cluttered environments a small quadrotor UAV might be better suited. 
For aerial manipulation a fully-actuated multirotor UAV might be the optimal candidate. 
Each of these UAV configurations has benefits and drawbacks in certain applications. 

\subsection{Literature Overview}
As applications for UAVs become more complex, with different requirements along their missions, morphable UAVs appeared. 
Systems of the class of morphable UAVs can change their configuration, optimizing the UAV's shape depending on a local task along the mission.
 
In \cite{falanga2018foldable} and \cite{riviere2018agile} aerial robots are presented that are able to translate the position of their propellers to squeeze through narrow gaps. 
For space-efficient storing and high speed ejection the quadrotor UAV in \cite{pastor2019design} has a body-drag optimized shape in folded configuration that unfolds for normal flight. 
In \cite{zhao2017whole, zhao2018design, zhao2017deformable} a snake-like multirotor platform is described, that can translate through air and grasp objects.

\begin{figure}[t]
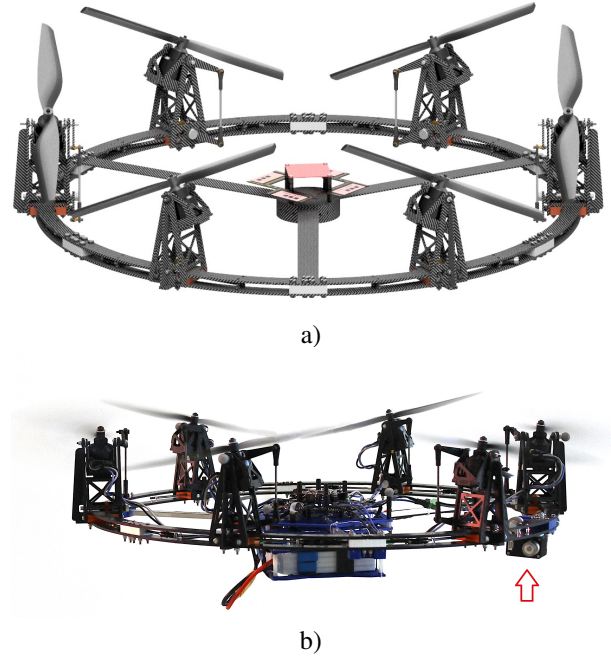

\centering
\figFASTHexProto\\ \centering a) \\
\figFASTHex\\ \centering b)
\caption{a) CAD prototype of the FAST-Hex. 
All propellers are tilted in a synchronized manner by a single motor. 
b) Flying prototype with tilted propellers. 
The single servomotor for tilting all propellers is visible on the right bottom side of the ring structure and highlighted with an arrow.}
\label{fig:FASTHexProto}
\end{figure}

A particular subset of morphable aerial robots achieve control of their body pose beyond the classical position and yaw orientation tracking.
The authors of \cite{kulkarni2019reconfigurable} present an aerial robot that can tilt a part of its frame in order to gain independent control of the vehicle's pitch angle, while the authors of \cite{8798092} lock the UAV's inner body in a gimbal system to achieve full pose tracking with the inner body. 
To allow full independent tracking of position and orientation trajectories the multirotor UAVs in \cite{2012-RylBueRob, 2013-RylBueRob, kamel2018voliro, InvernizziEtAl2018} can actively tilt all their propellers. 
This class of fully-actuated non-collinear multi-rotor systems has emerged as a class of UAVs benefiting from fast disturbance rejection~\cite{2014-VoyJia, 2013-SegAlrShySav, InvernizziLovera2018, InvernizziEtAl2020} and full-pose trajectory tracking (independent tracking of a desired 3D position and 3D orientation)~\cite{2011-CroLazMayLan, 2015-HuaHamMorSam, 2015-RylBueRob, 2015e-RajRylBueFra, 2013-LonWanCap, 2015-OosAbiNarKunKonUch, 2016-BreDan}). 
Furthermore, fully-actuated aerial vehicles are able to track a wrench profile (independent force and torque trajectories) making them optimal candidates as aerial-physical interaction tools. 

Technical solutions for fully-actuated aerial vehicles are currently implemented following two paradigms. 
Aerial vehicles of the first paradigm have their propellers fixed in a particular tilting angle (see our previous works~\cite{2015e-RajRylBueFra, 2019h-RylMusPieCatAntCacFra}) and do not belong to the group of morphable drones. 
These systems have simpler mechanics, lower control complexity and are usually lighter as no additional actuators are required, but suffer from increased energy consumption due to unavoidable, parasitic internal forces and a usually smaller volume of admissible wrench. 
Systems of the second paradigm can change the pose of the propellers, allowing thrust vectoring of every single propeller (cf.~\cite{2015-RylBueRob, kamel2018voliro, InvernizziEtAl2018}). 
While these systems commonly enable tracking of a larger or tunable volume of admissible wrench and therefore waste less energy, the mechanics and the control of these systems are more complex and the weight is increased by the number of required actuators, decreasing the overall flight time.

%Unmanned Aerial Vehicles (UAVs) are widely used in different application scenarios like remote monitoring and aerial photography, search and rescue missions and, more and more, aerial physical interaction with the environment~\cite{2014b-GioRylPraBueFra, 2014h-YueSecBueFra} and the with the human operator~\cite{2014e-SteBasBueFra}. 
%This new field of complex tasks including grasping and manipulation results in new challenges in the mechanical structure, the design and furthermore the control of aerial vehicles. 

%Recently, fully-actuated non-collinear multi-rotor systems emerged as a valid solution to either benefit from a faster disturbance rejection~\cite{2014-VoyJia, 2013-SegAlrShySav} or achieve a full-pose tracking, \ie a decoupled tracking of a desired 3D position and 3D orientation~\cite{2011-CroLazMayLan, 2015-HuaHamMorSam, 2015-RylBueRob, 2015e-RajRylBueFra, 2013-LonWanCap, 2015-OosAbiNarKunKonUch, 2016-BreDan}. 
%Furthermore, fully-actuated systems can be adapted to track a desired wrench and are therefore optimal tools in physical interaction tasks.
%The benefits come with the drawback of a reduced efficiency with respect to standard collinear multi-rotors due to higher internal forces.

\subsection{Contribution of this work}
In this article we present the Fully--Actuated by Synchronized--Tilting Hexarotor (\textit{FAST-Hex}), with six propellers actively tiltable by only one additional motor (see \figref{fig:FASTHexProto} a) \& b). 
This aerial platform allows wrench tracking in a large volume while using only one additional servomotor reducing the total mass, the probability of failure, energy consumption and complexity of the system.
The additional control input drives the configuration of the aerial platform in a continuum of configurations between the energetically very efficient but under-actuated configuration and the less efficient but maximally actuated configuration. 
By combining the best of the two worlds of under- and fully-actuated platforms, by means of only one additional servomotor we enable high-level fine tuning between maximal efficiency and decoupled wrench tracking for the task at hand.

This paper is an extension of work originally presented in~\cite{2016j-RylBicFra} and~\cite{2018d-FraCarBicRyl} where the theoretical idea of the FAST-Hex and an extension of the control concepts have been presented. 

The contribution of the paper is first, the presentation and discussion of the mechanics of the FAST-Hex prototype, that uses only one additional motor for actuating coordinately all propellers. 
The prototype overcomes the common star-form of multirotors by presenting a lightweight but rigid ring-structure. 
Second, we present an improved version of the pose-tracking controller presented in~\cite{2018d-FraCarBicRyl}, making it more suitable for such morphable platform. 
The pose tracking controller uses as input an arbitrary, desired full pose trajectory in $\mathbb{R}^3\times \mathrm{SO}(3)$ while the controller updates the orientation tracking, when strictly needed to overcome spinning rate saturations of any propeller. 
While this controller finds its perfect application in systems that can seamlessly transition between under and fully-actuated systems, it is applicable to any multi-rotor platform. 
The third contribution is a broad set of experiments conducted with the FAST-Hex prototype, demonstrating its superiority with respect to many other aerial platforms.

The paper is structured as follows. 
We first present the mechanical system of the FAST-Hex and then derive the dynamical model in Sec. \ref{sec:mech_design} and \ref{sec:simpl_model}. 
In \secref{sec:contrl} we describe the full-pose geometric control in $\mathbb{R}^3\times \mathrm{SO}(3)$ for generic multi-rotor platforms. 
In \secref{sec:exp} we present a broad spectrum of experimental results. 
Finally, \secref{sec:conclusion} concludes the paper with a summary of the results and an outline of future work.

%%%%%%%%%%%%%%%%%%%%%%%%%%%%%%%%%%%%%%%%%%%%%%%%%%%%%%%%%%%%%%%%%%%%%%
\section{System Design}\label{sec:mech_design}
%%%%%%%%%%%%%%%%%%%%%%%%%%%%%%%%%%%%%%%%%%%%%%%%%%%%%%%%%%%%%%%%%%%%%%
In this section we will describe the mechanical and electrical design of the FAST-Hex prototype.
\begin{figure}[t]
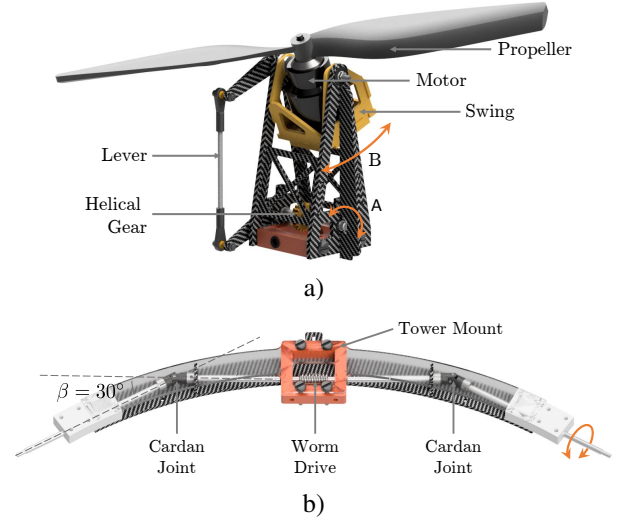

\centering
\figFASTHexTower\\ \centering a) \\
\figFASTHexRing\\ \centering b)
\caption{a) CAD model of a single motor tower. 
A worm drive actuates the helical gear indicated by `A' in the figure. 
The helical gear is rigidly connected on an axle, that is linked to a lever. 
The lever actuates a swing, which hosts the motor. 
The swing construction is used to rotate the the propeller close to its center. 
b) The complete MAV consists of six of the depicted elements. 
The top part of the ring is drawn transparent, allowing to see inside the ring structure. 
A single motor (not depicted in this figure) actuates the axes in the ring structure, that are connected with cardan joints. 
The direction of the worm drives is alternating, allowing the opposing rotation of neighboring motors.}
\label{fig:FASTHexSchemetic}
\end{figure}

\subsection{Mechanical Design}
\begin{figure}[t]
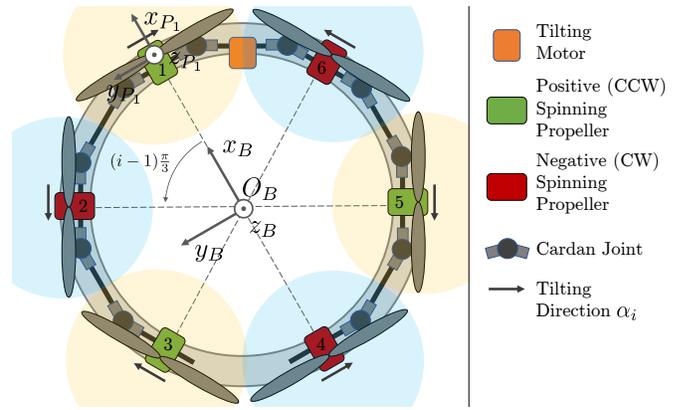

\centering
\figSchematic
\vspace{-2mm}
\caption{A sketch of the simplified model of FAST-Hex highlighting major mechanical components and the tilting directions of the six swings inside the motor towers. Counter-clockwise spinning propellers $\{1, 3, 5\}$ are depicted in light-orange, while the clockwise spinning ones $\{2, 4, 6\}$ in light-blue.}
\label{fig:Schematics}
\end{figure}

We designed a Micro Aerial Vehicle (MAV) that inherits the benefits of both, under- and fully-actuated vehicles, namely the possibility for energy efficient flight, \eg for cruising and the ability for independent position and orientation control, \eg for aerial manipulation or advance maneuvering in cluttered environments, while minimizing additional inputs, mechanical components and weight. 
Thanks to their simple mechanical design the most common fully-actuated MAVs are hexarotor systems composed by alternatingly fixed tilted propellers~\cite{2015e-RajRylBueFra, 2019h-RylMusPieCatAntCacFra}. 
These systems allow for full actuation in a limited state space of the MAV, depending on the tilt angle of the propellers. 
The larger the tilting angle, the more the platform is able to generate lateral forces but at the cost of higher internal forces, reducing the efficiency and flight-time of these platforms. 
The FAST-Hex is inspired by this MAV type. 
We aimed to be able to change the tilting angle while flying with a minimum set of additional inputs, namely only one additional actuator (see \figref{fig:FASTHexProto}-b). 
Therefore, the actuation of the single motor needs to be transmitted to all propellers (see \figref{fig:FASTHexSchemetic} and the attached video). To achieve this objective, all motors are aligned on a regular ring structure of radius $l$ (where  $l=\SI{0.305}{\metre}$ in  our prototype). 
The propellers are mounted on-top of six motor towers (see \figref{fig:FASTHexSchemetic}-a and Fig.\ref{fig:Schematics}), while all motor towers are evenly spaced on the ring planar structure and therefore $\SI{60}{\degree}$ apart. 
In order to simplify the motion model and minimizing the translation of the thrust generation points (\ie the center of the propellers) we aimed to rotate the propellers as close as possible to the rotation center of the blades. 
Therefore we designed the swing mechanism, rotating the propellers less than $\SI{1}{\centi\metre}$ away from their rotation centres (see \figref{fig:FASTHexSchemetic}-a).
The motors with the propellers are mounted in the swings in the top of the tower. They are rigidly connected via a lever mechanism to a worm drive with a high gear ratio (20:1) in the base of the tower. 
The worm drive offers self blocking capabilities, minimal play and precise control of the desired tilting angle. 
Inside the structural ring there are 11 carbon fibre axles, forming a polygon inscribed in the ring, all connected by Cardan joints (also known as universal joints): these allow the propagation of the rotation of the bars throughout the ring, see \figref{fig:FASTHexSchemetic}-b~\&~\ref{fig:Schematics}. The central axle is attached to a motor actuating the system. Consequently the propulsive groups 1-2-3 and 6-5-4 are actuated by two separate chains departing both from the same servo motor, a \mbox{Dynamixel MX-28T}, comprising a \mbox{Maxon} DC motor, a \mbox{ CORTEX-M3} micro-controller and a $\SI{12}{\bit}$ contactless encoder. 
Splitting the whole chain in two sub-chains greatly reduces friction phenomena and torsion effects of the carbon fibre parts, which, in the case of longer chains, could induce jerky movements on the parts located far from the motor box. 
Every second axles is endowed with the aforementioned worm drive (a worm-shaft coupled with a worm gear), that is responsible for the transmission of the rotation to the corresponding motor tower. The worm shafts and the gears are realized with a high-precision 3D-printer. The maximum absolute value of the tilting angle (mechanically limited) is $\overline{\alpha}=\SI{35}{\degree}$. 

Cardan joints have the well known property of an unequal input angle $\gamma_\mathrm{j-1}$ and output angle $\gamma_\mathrm{j}$ during a full rotation, depending on the bending angle $\beta$. As depicted in \figref{fig:Schematics}, there is one universal joint between the servo motor and the worm drive actuating propeller 1 and propeller 6, three universal joints to propeller 2 and propeller 5 and finally five joints to propeller 3 and propeller 4. 
To understand the effect size of this parasitic effects on the actual propeller tilting angles $\alpha_i$, we modeled the full drive train. Let us define $\gamma_j$ as the rotation angle of an axle placed downstream of a chain of $j$ previous universal joints. The actual propeller tilting angle $\alpha_\mathrm{i}$ depends on the desired tilting angle $\alpha_\mathrm{des}$, the transmission ratio $k$ of the worm drives and the propeller number (see Fig.~\ref{fig:Schematics}) and can be found in a recursive way ass
\begin{align}
\begin{split}
\gamma_0 &= \frac{1}{k} \, \alpha_\mathrm{des}, \\
\gamma_j &= \atantwo(\sin{\gamma_{i-1}},\cos{\beta}\cos{\gamma_{i-1}}) \quad j\in{[1,5]}, \\ 
\alpha_i &= k \, (-1)^{i-1} \, \gamma_{(6-|2i-7|)} \hspace{5em} i\in{[1,6]}.
\end{split}
\end{align}

\begin{figure}[t]
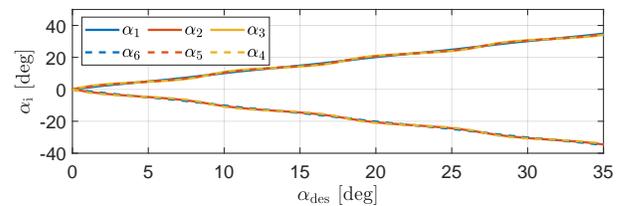

\centering 
\figCardJointAngle\\
\centering
%\figCardJointAngleError\\
\vspace{-2mm}
\caption{Desired $\alpha_\mathrm{des}$ versus actual tilting angle $\alpha_\mathrm{i}$ for the FAST-Hex for all six propellers. The absolute peak divergence is approximately $\SI{1}{\degree}$. Therefore it has been decided not to consider it in the control design but to treat it as disturbance.}
\label{fig:CardAngle}
\end{figure}

A comparison of the desired and the actual angles is depicted in \figref{fig:CardAngle}. 
The worm drives, with a transmission ratio of $k=0.05$, reduce the parasitic effect.
It becomes obvious that the maximum tracking difference for the two propellers with the most Cardan joints in between (propeller 3 and propeller 4) is approximately $\SI{1}{\degree}$. 
We will therefore neglect this relatively small difference and will let the controller (Sec.~\ref{sec:contrl}) cope with it. The overall structure of the ring gives a high rigidity to the system, reducing the vibrations of the motors, compared to the typical arm structure of multi-rotor systems. Mechanical details of the system are listed in Table \ref{tab:mechParams}.

%\subsection{Electro-Mechanical Components}\label{sec:real}
The electronics, including inertial measurement unit (IMU) and brushless motor controllers, are mounted in the center of the ring structure, decoupling parasitic vibrations from the motors. The IMU and the motor controllers are available off-the-shelf from Mikrokopter\footnote{https://www.mikrocontroller.com/}. 
The hardware is composed by $6$ \mbox{MK3638} motors, controlled by $6$ \mbox{BL-Ctrl V2.0} brushless controllers, and driving $6$ \mbox{EPP1245} propellers ($\SI{12}{inch}$ of diameter and $\SI{4.5}{inch}$ of pitch). 
The electronic speed controllers allow to precisely control the propeller spinning velocity using a closed loop sliding-mode controller~\cite{2017c-FraMal}.
The speed controllers are connected to a \mbox{Flight-Ctrl~V2.5} board, equipped with the IMU using 3 \mbox{ADXRS620} gyroscopes and a \mbox{Memsic MXR9500M} 3D accelerometer.

\begin{table}[]
\centering
\vspace{2mm}
\caption{Mechanical, physical and control parameters}
\renewcommand\arraystretch{1.2}
\resizebox{\columnwidth}{!}{
\begin{tabular}{lll}
\hline
Part                       	& Symbol   				& Value	\\ \hline
Ring ext. diameter          & $d$        			& $\SI{640}{\milli\metre}$	\\
Propeller diameter      	& 	  					& $\SI{12}{inch}$ ($\approx\SI{30.5}{\centi\meter}$)	\\
Propeller tilting angle 	& $\alpha_i$			& $(-1)^{i-1}\ |\alpha|$	\\
Tilting angle range           & $|\alpha|$ 			& $\in [\ang{0},\ \ang{35}]$	\\ 
Max tilting velocity       	& $\overline{\dot{\alpha}}=-\underline{\dot{\alpha}}$ & $\SI{10}{\degree\per\second}$	\\
Total mass   				& $m$        			& $\SI{3.1}{\kilo\gram}$	\\
Total inertia 				& $\Jm(i,i)\rvert_{i=1,2,3}$	&  $[0.089 \ 0.091 \ 0.164]^{\top}$ $\si{\kilo\gram\metre\squared}$	\\
Max propeller spin			& $\overline{w_i}$ 		& $\SI{102}{\hertz}$	\\
Min propeller spin 			& $\underline{w_i}$ 	& $\SI{16}{\hertz}$	\\
Max propeller force			& $\overline{f}_i$  	& $\SI{10} {\newton}$ \\
Max lift force          	& $f_{z,\mathrm{max}}$  & $\SI{60}{\newton}$	\\
Max lateral force       	& $f_{xy,\mathrm{max}}$ & $\SI{6}{\newton}$	\\
%Propeller direction 	&			 			& \red{$(-1)^{i-1}\ [\ ]$}	\\
Thrust coefficient 			& $c_f$					& $\SI{9.9e-4}{\newton\per\hertz\squared}$	\\
Drag-moment coefficient 			& $c_f^{\tau}$ 			& $\SI{1.9e-2}{\metre}$	\\
\hline
Propeller attitude 				& $\Rm_{A_i}^B $		& $\Rm_z \big ((i-1)\frac{\pi}{3})\big)\Rm_x(\alpha_i)\Rm_y(\beta)$	\\
\ith Propeller position		& $\pv_{A_i}^B$ 		& $\Rm_z \big((i-1)\frac{\pi}{3})\big) [\ell\ 0\ 0]^{\top}$	\\
\hline	
Proportional gain (trasl.) & $\Km_p(j,j)|_{j=1,2,3}$ & 50, 50, 50 [\ ] \\
Integral gain (trasl.) & $\Km_{p_i}(j,j)|_{j=1,2,3}$ & 20, 20, 20 [\ ] \\
Derivative gain (trasl.) & $\Km_v(j,j)|_{j=1,2,3}$ & 14.14, 14.14, 14.14 [\ ] \\
Proportional gain (rot.) & $\Km_R(j,j)|_{j=1,2,3}$ & 15, 15, 6 [\ ] \\
Integral gain (rot.) & $\Km_{R_i}(j,j)|_{j=1,2,3}$ & 1, 1, 1 [\ ] \\
Derivative gain (rot.) & $\Km_w(j,j)|_{j=1,2,3}$ & 1.5, 1.5, 0.5 [\ ] \\
%\hline
%& $\overline{\ev}_{p_i}=-\underline{\ev}_{p_i}$ & $[3\ 3\ 3]^{\top}$ \si{\meter} \\
%& $\overline{\ev}_{R_i}=-\underline{\ev}_{R_i}$ & $[1\ 1\ 1]^{\top}$ \si{rad} \\
%\hline
%& $n_{\text{it}}$ & 20  [\ ] \\
%& $\overline{w_i}$ & 98 \si{Hz} \\
%& $\underline{w_i}$ & 20 \si{Hz} \\
%& $\overline{\dot{\alpha}}=-\underline{\dot{\alpha}}$ & 10 \si{deg\per\second} \\
\hline
\end{tabular}
\label{tab:mechParams}
}
\end{table}

%%%%%%%%%%%%%%%%%%%%%%%%%%%%%%%%%%%%%%%%%%%%%%%%%%%%%%%%%%%%%%%%%%%%%%
\section{Modeling}\label{sec:simpl_model}
%%%%%%%%%%%%%%%%%%%%%%%%%%%%%%%%%%%%%%%%%%%%%%%%%%%%%%%%%%%%%%%%%%%%%%
A photograph and a CAD model of the actual FAST-Hex are shown in \figref{fig:FASTHexProto}. 
We will now introduce a simplified mathematical model of the FAST-Hex that we will utilize deriving the controller in \secref{sec:contrl}. 
A sketch of the simplified model is depicted in \figref{fig:Schematics} showing the relevant reference frames. 
This simplified model has been introduced in~\cite{2016j-RylBicFra} - we will therefore only summarize it here.

The simplified FAST-Hex model is composed by a rigid body and six mass-free and orientable propellers. 
We define a world frame $\wFrame=O_W,\{ \vx_W, \vy_W, \vz_W\}$ and a body frame $\bFrame=O_B,\{ \vx_B, \vy_B, \vz_B\}$ that is rigidly attached to the FAST-Hex with $O_B$ being the geometric center and the center of mass (CoM) of the system (see \figref{fig:FASTHexProto}). 
The position of $O_B$ is represented in $\wFrame$ by denoting $\vect{p}_B \in \nR{3}$ and the attitude of $\bFrame$ in $\wFrame$ is expressed by the rotation matrix $\rotMatB \in \mathrm{SO}(3)$. 
The angular velocity of the body frame $\bFrame$ with respect to the world frame $\wFrame$ represented in $\bFrame$ is denoted with $\vomegaB \in \nR{3}$. 
The attitude kinematics of the body $\rotMatB$ is then given by 
\begin{align}
\dot{\vect{R}}_B = \vect{R}_B [{\bs\omega}_B]_{\times},
\end{align}
where $[\bullet]_{\times}\in {\mathrm{so}}(3)$ represents any skew symmetric matrix associated to any vector $\bullet \in \mathbb{R}^3$.

%\begin{figure}[t]
%\centering
%\figSimpleModel
%\caption{A sketch of the simplified model of FAST-Hex. 
%We highlighted all quantities related to the ($i$$=$$3$)-th propeller group. 
%Counter-clockwise spinning propellers $\{1, 3, 5\}$ are depicted in light-orange, while the clockwise spinning ones $\{2, 4, 6\}$ in light-blue.}
%\label{fig:SimplifiedModel}
%\vspace{-0.5cm}
%\end{figure} 

Next we introduce the six propeller frames ${\cal F}_{P_1},\ldots, {\cal F}_{P_6}$ with ${\cal F}_{P_i}=O_{P_i},\{ \vx_{P_i}, \vy_{P_i}, \vz_{P_i}\}$. 
We denote $\vect{e}_1$, $\vect{e}_2$, and $\vect{e}_3$  the three vectors of the canonical basis of $\mathbb{R}^3$, and with $\vect{R}_x$ and $\vect{R}_z$ the two canonical rotation matrices in $\mathrm{SO}(3)$. 
The orientation of the \ith propeller ${\cal F}_{P_i}$ can now be expressed with respect to body frame ${\cal F}_B$ by the rotation matrix 
\begin{align}
\vect{R}_{P_i}^B(\alpha) = \mathbf{R}_z\left((i-1)\frac{\pi}{3}\right) \mathbf{R}_x \Big((-1)^{i-1}\alpha \Big), \quad i=1,\ldots, 6 \label{eq:RSi_def}
\end{align}
where $\alpha\in\mathcal{A}$ is the \emph{synchronized tilting angle} which is adjustable by using the single servomotor (see \figref{fig:FASTHexProto}). The presence of $(-1)^{i-1}$ in~\eqref{eq:RSi_def} represents the effect that propellers with adjacent indexes are \emph{tilting} in opposite directions, which guarantees the full actuation of the platform for $\alpha\in\mathcal{A}\backslash \{0\}$, see, \eg~\cite{2015e-RajRylBueFra, 2018a-MicRylFra} for more details on the design of fully actuated platforms.

The vector originating from $O_B$ to $O_{P_i}$, representing the position of the center of the \ith propeller, expressed in body frame $\bFrame$, is 
\begin{align}
\vect{p}^B_{B,P_i}= l\mathbf{R}_z\left((i-1)\frac{\pi}{3}\right) \vect{e}_1,
\quad \text{for } i=1,\ldots, 6 
\end{align}
with $l>0$ being the distance from $O_B$ to $O_{P_i}$. 
The six propellers are centered in $O_{P_i}$ and spin with the angular velocity $(-1)^{i-1}w_i\vz_{P_i}$, where $(-1)^i$ models the property that propellers with adjacent indexes are designed to \emph{spin} with opposite sign and therefore generate opposite drag torques. The six propeller spinning rates $w_i>0$ are as usual individually controllable. 

In the following we derive the dynamics of motion of the FAST-Hex platform which is actuated by changing the spinning velocity and synchronized orientation of the six propellers. 
While spinning, the propellers generate in a sufficient approximation a thrust force $\vect{f}_i$ and a drag moment $\boldsymbol{\tau}_i$, applied in $O_{P_i}$ and oriented along $\vz_{P_i}$, which are expressed in $\bFrame$ as%

\begin{align}
\vect{f}_i^B(f_i,\alpha) &= \vect{R}^B_{P_i}(\alpha)\vect{f}_i, \quad \text{for } i=1,\ldots, 6, \quad\text{and} \label{eq:propell_force} \\
\boldsymbol{\tau}_i^B(f_i,\alpha) &= (-1)^{i} c_f^\tau \vect{R}^B_{P_i}(\alpha) \vect{f}_i,	 
\quad \text{for } i=1,\ldots, 6. 	 \label{eq:propell_drag} 
\end{align}

In \eqref{eq:propell_force} $c_f^\tau>0$ is a constant parameter characterizing the relationship of the generated force and torque, depending on the physical parameters of the propeller. The scalar 
$f_i$ is the intensity of the force produced by the propeller, which is related to the controllable spinning rate $w_i$ by means of the quadratic relation
\begin{align}
\vect{f}_i = c_f w_i^2 \vect{e}_3,
\end{align} 
where $c_f>0$ is another propeller shape dependent constant parameter. 

By summing all thrust forces we can find the total force applied to the FAST-Hex's CoM, expressed in world frame $\wFrame$ as%
\begin{align}
%\vect{f}^W=
\vect{f}^W(\alpha,\vect{u}) = \vect{R}_B\sum_{i=1}^6\vect{f}_i^B(f_i,\alpha) = \vect{R}_B\vect{F}_1(\alpha) \vect{u},\label{eq:total_force}
\end{align}
where $ \vect{u}=[f_1\;f_2\;f_3\;f_4\;f_5\;f_6]^{\top}$
%\red{AF: or we can chose $ \vect{u}=[w^2_1\;w^2_2\;w^2_3\;w^2_4\;w^2_5\;w^2_6]^{\top}$}
 and
$\vect{F}_1(\alpha)\in\mathbb{R}^{3\times 6}$ is a suitable $\alpha$-dependent matrix.
% that
% maps $\vect{u}$ to the total force applied to the platform expressed in body frame.
For the case $\alpha=0$ all propellers are collinear (as for a standard hexarotor), then $\vect{F}_1(\alpha=0)= [\vect{0}_6^\top\;\vect{0}_6^\top\;\vect{1}_6^\top]^{\top}$.

By adding all torque contributions, namely the drag moments \eqref{eq:propell_drag} and the thrust contributions \eqref{eq:propell_force}, we compute the total moment applied to the platform's CoM, with respect to $O_B$, and expressed in $\bFrame$ as
\begin{align}
\begin{split}
{\bs \tau}^B(\alpha,\vect{u}) =& \sum_{i=1}^6\left(\left(\vect{p}^B_{B,P_i}\times \vect{f}_i^B(f_i,\alpha)\right) + {\bs\tau}_i^B(f_i,\alpha) \right) \\
=&\ \vect{F}_2(\alpha) \vect{u}.\label{eq:total_moment}
\end{split}
\end{align}

The equations of motion of the aerial platform can be compactly expressed by using the Newton-Euler approach
\begin{align}
\begin{bmatrix}
m\ddot{\vect{p}}_B \\
\vect{J}\dot{{\bs \omega}}_B
\end{bmatrix}
= 
%\begin{bmatrix}
%m \vect{I}_3 & \bs 0\\
%\bs 0 & \vect{J}
%\end{bmatrix}^{-1}
%\left(
%-
-\begin{bmatrix}
m g \vect{e}_3\\
{\bs \omega}_B \times \vect{J}{\bs \omega}_B
\end{bmatrix}
%\right),
+
\begin{bmatrix}
\vect{f}^W\\
{\bs \tau}^B
\end{bmatrix}
\label{eq:newt-eul}
\end{align}
where $\vect{J} > 0$ represents the $3\times 3$ inertia matrix of the rigid body with respect to $O_B$ and expressed in $\bFrame$, $m>0$ represents the total mass of the FAST-Hex, and finally $g>0$ is the gravitational acceleration.

Replacing~\eqref{eq:total_force} and~\eqref{eq:total_moment} in~\eqref{eq:newt-eul} we obtain 
\begin{align}
\begin{bmatrix}
m\ddot{\vect{p}}_B \\
\vect{J}\dot{{\bs \omega}}_B
\end{bmatrix}
=
-\begin{bmatrix}
m g \vect{e}_3\\
{\bs \omega}_B \times \vect{J}{\bs \omega}_B
\end{bmatrix}
+
\underbrace{
\begin{bmatrix}
\vect{R}_B\vect{F}_1(\alpha)\\
\vect{F}_2(\alpha)
\end{bmatrix}
}_{\vect{F}(\vect{R}_B,\alpha)}
\vect{u}.
\label{eq:newt-eul-2}
\end{align}

Finally, we will take propeller spinning rate saturations into account, which can be expressed as input limits as
\begin{align}
\vect{u} \in {\cal U} = \{ \vect{u}\in \mathbb{R}^6\,|\,0\leq\underline{u}\leq f_i \leq\overline{u} \quad \forall i=1\ldots 6\}.
\label{eq:input_constraint}
\end{align}
where $\underline{u}\approx 0^+$ is the lower and $\overline{u}$ are related to the upper spinning rate limit. 
While the upper spinning rate limit has obvious actuator reasons, we additionally introduce a lower spinning rate limit as efficient propellers are optimized for a particular spinning direction and most propeller-motor controllers use an open loop propeller starting procedure with an undefined starting time making stopping undesirable~\cite{2017c-FraMal}.

\subsection{Discussion on model simplifications}\label{sec:model_discussion}
The presented, simplified FAST-Hex model neglects several properties of the actual system. 
In the following we list the unmodeled properties and comment on their impact. 
While actively tilting the propellers, the gyroscopic effect causes a torque, perpendicular to the angular momentum of the propellers and the tilting direction. 
This gyroscopic effect is small due to the low mass of the propellers and the slow tilting velocity ($\overline{\alpha} = \SI{10}{\degree\per\second}$) and we therefore neglect it. 
For the same reason, we ignore the multi-body dynamics between the actuated propellers and the main body. 
The actuation of the propellers causes a position change of the CoM and a change of the inertia matrix $J$ of the main body in~\eqref{eq:newt-eul}. 
These changes are as well small ($\Delta \vect{p}^B_{B,P_i} < \SI{0.5}{\percent}$ in \eqref{eq:total_moment}). 
Additionally, we neglect the effects of the universal joints and the resulting minor position change of the propellers due to the actuation. 

This work focuses on the mechanical design and the control of the FAST-Hex under a low velocity flight regime. 
We will therefore neglect aerodynamic effects such as the well-known first-order effects rotor drag, fuselage drag, and H-force, as these effects depend linearly on the vehicle's velocity and can therefore be neglected att small velocities \cite{2018c-FaeFraSca}.

We will demonstrate in the experimental results section (see \secref{sec:exp}) that the controller presented in \secref{sec:contrl} can sufficiently cope with these uncertainties. 

\subsection{Synchronized Tilting Angle: Efficiency vs. Full-Actuation}\label{sec:alpha_discussion}
The FAST-Hex, with the tilting angle being $\alpha\in[\ang{0}\ \ang{35}]$, has two structurally different configurations:
\begin{enumerate}
\item $\alpha=0$ \; $\Rightarrow$ \; $\rank\big(\vect{F}(\vect{R}_B,\alpha=0)\big)=4$
\item $ \alpha \in \mathcal{A}\backslash \{0\}$ \; $\Rightarrow$ \; $\rank\big(\vect{F}(\vect{R}_B,\alpha)\big)=6$.
\end{enumerate}
In case the FAST-Hex would allow for $\alpha<\ang{0}$ the system would have an additional rank loss at $\alpha = \ang{-3.56}$, in fact $\rank\big(\vect{F}(\vect{R}_B,\alpha = \ang{-3.56})\big)=5$ due to a yaw torque controllability loss \cite{2018r-MorBicRylFra}. We therefore restrict the tilting angle to positive values.

In configuration~1) all propellers of the FAST-Hex have collinear spinning axes.
We will therefore call this configuration \textit{\gls{udt}} configuration opposing the \textit{\gls{mdt}} in configuration~2). 
In UDT-configuration the system degenerates to an ordinary hexarotor platform. 
The internal forces in UDT-configuration are zero and only internal torques due to the drag moment appear. 
The internal torques due to drag moment are typically one order of magnitude less strong than the torques generated by the thrust moments and are therefore neglected in the following efficiency considerations.

We model the wasted (internal) force using the following index
\begin{align} 
\eta_f(\alpha,\vect{u}) = \tfrac{\|\sum_{i=1}^{6}\vect{f}_i^B(f_i,\alpha)\|}{\sum_{i=1}^{6}\|\vect{f}_i^B(f_i,\alpha)\|} 
=
\tfrac{\|\sum_{i=1}^{6}\vect{f}_i^B(f_i,\alpha)\|}{\sum_{i=1}^{6}f_i} 
\in [0,1]
\label{eq:force_efficiency}
\end{align}
that we call the \emph{force efficiency} index. 
It is easy to check that $\eta_f(\alpha=0,\vect{u})=1$ for any input $\vect{u}$, which corresponds to maximum efficiency. 
Hence the UDT-configuration is energetically very efficient. 
This comes with the drawback that the platform is under-actuated and a simultaneous tracking of fully independent $\vect{p}_r(t)$ and $\vect{R}_r(t)$ is impossible. 
The best choice left in this case is a control that selects a new reference orientation, denoted with $\vect{R}_d(t)$, that is compatible\footnote{Compatibility is related to the well-known differential flatness property of collinear-rotor vehicles. 
In particular, the $\vect{z}_B$ axis must be kept parallel to $\ddot{\vect{p}}_r(t) + mg \vect{e}_3$. 
The orientation about $\vect{z}_B$ is instead not constrained by $\vect{p}_r(t)$.} with $\vect{p}_r(t)$ and is as close a possible to $\vect{R}_r(t)$ with respect to a certain criterion, as, \eg possessing the same yaw angle of $\vect{R}_r(t)$, or the same projection of a certain axis on a certain plane. This approach is used, \eg by the well established geometric control~\cite{2010-LeeLeoMcc}, whose rotational part is based on~\cite{2006-MahChaHam}.
%, and has been implemented in many real robotics setups showing agile maneuvering, as, \eg in~\cite{2011-MelKum}.
Almost global convergence is achieved without the singularities of other orientation parametrization.

\begin{figure}[t]
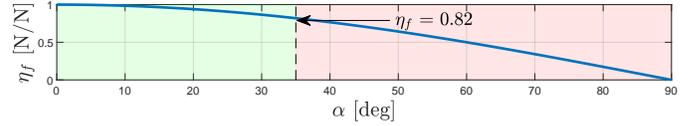

\centering
\figEfficiency
\vspace{-3mm}
\caption{Nominal efficiency of the FAST-Hex depending on the tilting angle $\alpha$ based on the efficiency index presented in \eqref{eq:force_efficiency}. 
The index is computed for horizontal hovering ($\vect{R}_B = I_{3\times 3}$) condition. 
For the maximum tilting angle $\alpha = \SI{35}{\degree}$ the efficiency drops to $0.82$, meaning that $\SI{18}{\percent}$ of the generated forces are wasted as internal forces.}
\label{fig:Efficiency}
\end{figure}

\begin{figure}[t]
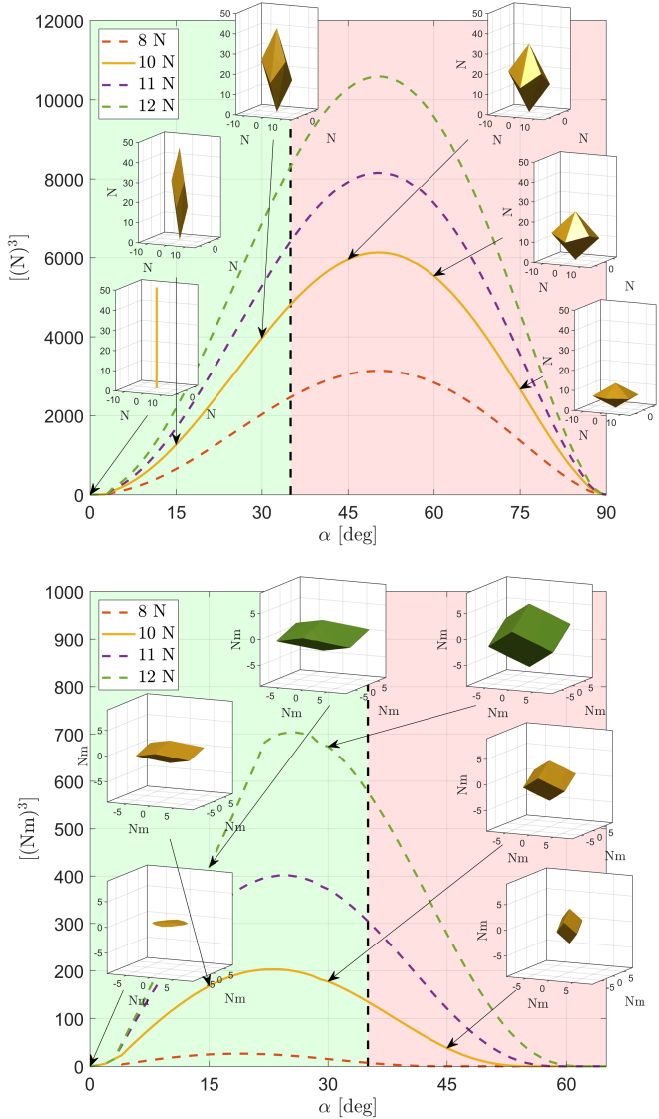

\centering
\figForceSpace\\\vspace{-1.5mm}
\figTorqueSpace
\vspace{-1em}
\caption{Top: Volume of attainable total forces $\vect{R}_B^{\top}\vect{f}^W(\alpha,\vect{u})$ corresponding to different values of $\alpha$.
The volumes are computed using~\eqref{eq:total_force}, expressed in the body frame ${\cal F}_B$, and imposing $\underline{w_i}\leq w_i \leq \overline{w_i}~\forall i=1, \ldots, 6$ and $\boldsymbol{\tau}^B = \vect{0}$. 
The larger $\alpha$ (inside the feasible set) the larger the volume of the polyhedron. 
For $\alpha=0$ the polyhedron degenerates to a single direction along the $\vect{z}_B$ axis. 
The different lines represent different limits for the maximum rotor spinning velocity $\overline{w_i}$. 
Bottom: Volume of attainable total torques $\boldsymbol{\tau}^B(\alpha,\vect{u})$ corresponding to different values of $\alpha$. 
The volumes are computed using~\eqref{eq:newt-eul-2}, expressed in the body frame ${\cal F}_B$, and imposing $\underline{w_i}\leq w_i \leq \overline{w_i}~\forall i=1, \ldots, 6$ and $\vect{f}^B = [0~0~mg]^{\top}$.}
\label{fig:ForceSpace}
\end{figure}

In MDT-configuration the internal forces in hovering are greater than zero, which means that the system is wasting more energy than in UDT-configuration. 
The larger $|\alpha|$ the larger the internal forces. 
This is clearly visible from the fact that $\eta_f(\alpha\in\mathcal{A}\backslash \{0\},\vect{u})< 1$. 
In particular, during horizontal hovering, when all the propellers are spinning at the same speed, producing the same force $f$, we have that $\eta_f(\alpha,f\vect{1}_{6\times 1})=\cos \alpha$. 
For horizontal hovering we plot the efficiency index in \figref{fig:Efficiency} for a changing tilting angle, showing that the efficiency drops to $\eta_f=0.82$ for maximum tilting of $\alpha=\SI{35}{\degree}$. 
If the platform is following a non-hovering trajectory then $\eta_f(\alpha,\vect{u})$ is in general different from $\cos \alpha$ and one has to use~\eqref{eq:force_efficiency} to exactly compute it.
On the other side in MDT-configurations the platform is fully-actuated, and the larger $|\alpha|$ the larger the volume of admissible total forces $\vect{f}^W$ in~\eqref{eq:newt-eul} as it can be clearly seen from \figref{fig:ForceSpace}. 
The simultaneous tracking of $\vect{p}_r(t)$ and $ \vect{R}_r(t)$ becomes feasible as shown in~\cite{2015e-RajRylBueFra}, where a controller for this particular case is also proposed.
We compared in \figref{fig:ForceSpace} the influence of the tilting angle and the actuator limitations $\overline{w_i},\underline{w_i}$ (see the limits in the inputs~\eqref{eq:input_constraint}) on the volume of admissible forces and torques depending on the tilting angle $\alpha$.
For computing the volume of admissible forces (top plot), we set the torques in~\eqref{eq:newt-eul-2} to $\boldsymbol{\tau}^B = \vect{0}\ \SI{}{\newton\metre}$, while for computing the volume of admissible torques (bottom plot) we set the forces to obtain $\vect{f} = [0\ 0\ mg]^{\top}\ \SI{}{\newton}$.
The results of these plots as well drove the decision to limit the tilting angle $\bar\alpha$ to maximum $\SI{35}{\degree}$ as the combined maximum torque and force volume is achieved at $\approx\SI{35}{\degree}$.

Due to the fact that $\alpha$ is a slowly changeable parameter, the change of $\alpha$ is delegated to a high-level slow-rate controller/planner or to a human operator. 
The high-level controller can gently tune $\alpha$ while flying, thus continuously changing the platform between configuration~1) and any of the configurations of type~2) in order to adapt to the particular task being executed. 
For example configuration~1) can be chosen when a pure horizontal hovering is requested while a type~2) configuration can be selected when hovering with non-zero roll and pitch is needed. 

%%%%%%%%%%%%%%%%%%%%%%%%%%%%%%%%%%%%%%%%%%%%%%%%%%%%%%%%%%%%%%%%%%%%%%
\section{Full-Pose Geometric Control with Prioritized Position Tracking}\label{sec:contrl}
%%%%%%%%%%%%%%%%%%%%%%%%%%%%%%%%%%%%%%%%%%%%%%%%%%%%%%%%%%%%%%%%%%%%%%
In this section, we present a control law for the six force inputs $\vect{u}$ in \eqref{eq:input_constraint} that lets $\vect{p}_B$ and $\vect{R}_B$ track at best an arbitrary full-pose reference trajectory $(\vect{p}_r(t), \vect{R}_r(t)): \mathbb{R} \to \nR{3} \times \mathrm{SO}(3)$. 
The time-varying parameter $\alpha$ is given to the controller. 
By decoupling the control of $\alpha$ and $\vect{u}$, we make the control law directly applicable for a broad spectrum of aerial vehicles beyond the scope of the FAST-Hex.

The most obvious approach to control the FAST-Hex would be to use the geometric controller presented in~\cite{2010-LeeLeoMcc} while in configuration~1) and the fully-actuated controller~\cite{2015e-RajRylBueFra} while in configuration~2). 
The first drawback of this approach concerns the challenges that might arise from switching between two controllers and the second is an ill-conditioned computation of $\vect{F}(\vect{R}_B,\alpha)^{-1}$ (used in~\cite{2015e-RajRylBueFra}) for $\alpha \rightarrow 0$. 
A possible solution to the ill-conditioned inversion would be to use the geometric controller~\cite{2010-LeeLeoMcc} for even small angles of $|\alpha|$, which would require abandoning full-pose tracking for small values of $\alpha$. 
However, it might be actually desirable to drive the FAST-Hex with a small $\alpha$ angle in order to find a trade-off between full actuation and minimization of wasted internal forces.

Therefore, we suggest using a control that works seamlessly in both configurations, an extension of the under-actuated geometric control~\cite{2010-LeeLeoMcc} for fully actuated platforms. 
The desired behavior of a platform driven by the controller will then be:
\begin{itemize}
\item The larger $\alpha$ the more the platform can realize an arbitrary force vector and track simultaneously a position and orientation trajectory. 
The FAST-Hex becomes gradually fully actuated.
\item The smaller $\alpha$ the more the output of the control law resembles~\cite{2010-LeeLeoMcc}. 
In other words, when $|\alpha|$ decreases the FAST-Hex becomes gradually under-actuated, \ie it still keeps a good tracking of the reference position but it becomes progressively unable to independently track also a generic reference orientation. 
\end{itemize}

%\item the smaller $|\alpha|$ (and the larger $\|\ddot{\vect{p}}_r(t)\|$) the more the output of the control law resembles~\cite{2010-LeeLeoMcc}. In other words, when $|\alpha|$ decreases FAST-Hex becomes gradually under-actuated, \ie it still keeps a good tracking of the reference position but it becomes progressively unable to independently track also the reference orientation; 
%\item the larger $|\alpha|$ (and the smaller $\|\ddot{\vect{p}}_r(t)\|$) the 
% more the controller generalizes~\cite{2010-LeeLeoMcc}, and FAST-Hex becomes gradually fully-actuated, \ie more and more able to control simultaneously the position and the orientation in an independent way.

The implemented controller is an improvement of the full-pose geometric controller with prioritised position tracking described in \cite{2016j-RylBicFra} which is composed by an inner \textit{attitude controller} and an outer \textit{position controller}.
The controllers are then cascaded by a \textit{wrench mapper} which computes the actuators set-point $\textbf{u}$ according to the desired control force ${\textbf{u}}_f \in \mathbb{R}^3$ and moment $\textbf{u}_{\tau} \in \mathbb{R}^3$ provided by the position and attitude controllers, respectively.
In the following the three components are described  in detail.

\subsection{Position control}\label{subsec:ctrl:pos}
The position controller takes as input the full-pose trajectory ($\textbf{p}_r$, $\dot{\textbf{p}}_r$, $\ddot{\textbf{p}}_r \in \mathbb{R}^3$ and $\textbf{R}_r = [\textbf{b}_{1r} \textbf{b}_{2r} \textbf{b}_{3r}]\in \mathrm{SO}(3)$), the measured position ${\textbf{p}}_B$, the measured linear velocity ${\dot{\textbf{p}}}_B$ and the measured attitude $\textbf{R}_B$.
It produces as output the desired orientation $\textbf{R}_d \in \mathrm{SO}(3)$ and the desired control force $\textbf{u}_f$.

\subsubsection{Control equations}\label{subsubsec:ctrl:pos:equ}
Given the considered input one can define the position and velocity tracking errors respectively as follows:
\begin{align}
\textbf{e}_p = \textbf{p}_B - \textbf{p}_r, & & \textbf{e}_v = \dot{\textbf{e}}_p = \dot{\textbf{p}}_B - \dot{\textbf{p}}_r.
\end{align}
It is then possible to define the integral position tracking error as
\begin{equation}
\textbf{e}_{pi} = \int_{0}^{t} \textbf{e}_p d\tau.
\end{equation}

The reference force vector is then computed as
\begin{equation}
\textbf{f}_r = m\left(\ddot{\textbf{p}}_r + g\textbf{e}_3\right) - \textbf{K}_p \textbf{e}_p - \textbf{K}_{pi} \textbf{e}_{pi} - \textbf{K}_v \textbf{e}_v,
\end{equation}
where $\textbf{K}_p$, $\textbf{K}_{pi}$, $\textbf{K}_v \in \mathbb{R}^{3\times3}$ are positive diagonal matrices.

Such force vector is then rotated from the inertial to the body frame and saturated assuming a cylindric bounded force as described in \cite{2018d-FraCarBicRyl} in order to obtain the desired control force
\begin{align}\label{eq:ctrlForce}
&\textbf{u}_f = sat_{\mathcal{U}_{xy}}\left(({\textbf{f}_r}^{\top} \textbf{R}_B \textbf{e}_1) \textbf{e}_1 + ({\textbf{f}_r}^{\top} \textbf{R}_B \textbf{e}_2) \textbf{e}_2 \right) + ({\textbf{f}_r}^{\top} \textbf{R}_B \textbf{e}_3) \textbf{e}_3, \\
&\mathcal{U}_{xy} \left( \alpha \right) = \big\{ \begin{bmatrix}u_1 & u_2 \end{bmatrix}^{\top} \in \mathbb{R}^2 | u_1^2 + u_2^2 \leq r_{xy}^2 \left( \alpha \right) \big\},
\end{align}
where $r_{xy}\left( \alpha \right)$ will be described later.

The desired orientation, instead, is computed taking into account the requested orientation, the reference force vector, and the lateral force bound as described in Algorithm \ref{algo:bisection}.
In particular, $c_{\theta}$ and $s_{\theta}$ are the cosine and sine of $\theta$ respectively.
Finally, it is possible to compute the desired orientation as

\begin{align}
\textbf{b}_{3d} &= \textbf{b}_{3r}c_{\theta} + \left( \textbf{k} \times \textbf{b}_{3r} \right) s_{\theta} + \textbf{k} \left( \textbf{k} \cdot \textbf{b}_{3r}\right)\left( 1 - c_{\theta}\right)\\
\textbf{R}_d &= 
\begin{bmatrix} 
\underbrace{\left( \textbf{b}_{3d} \times \textbf{b}_{1r} \right) \times \textbf{b}_{3d}}_{\textbf{b}_{1d}} & \underbrace{\textbf{b}_{3d} \times \textbf{b}_{1r}}_{\textbf{b}_{2d}} & \textbf{b}_{3d}
\end{bmatrix}.
\end{align}

\begin{algorithm}[tb]
\setlength{\textfloatsep}{0pt}% Remove \textfloatsep
\label{algo:bisection}
\caption{Computation of $\textbf{R}_d$ via bisection method}
\small{
\KwData{$n_{\text{it}}$ (number of iterations $\propto$ solution accuracy)}
\KwData{$\textbf{b}_{3r}$, $\textbf{f}_r$, $r_{xy}(\alpha)$}
%%%
$\theta_{max}\gets \arcsin\left(\frac{\left\lVert \textbf{b}_{3r}\times \textbf{f}_r \right\lVert}{\left\lVert \textbf{f}_r \right\lVert}\right)$, $\theta\gets \frac{\theta_{max}}{2}$, $\textbf{k}\gets \frac{\textbf{b}_{3r}\times \textbf{f}_r}{\left\lVert \textbf{b}_{3r}\times \textbf{f}_r \right\lVert}$\; 
%%%
\For{$i=1$ to $n_{\text{it}}$}{
	$\textbf{b}_{3d} \gets \textbf{b}_{3r}c_{\theta} + \left( \textbf{k} \times \textbf{b}_{3r} \right) s_{\theta} + \textbf{k} \left( \textbf{k} \cdot \textbf{b}_{3r}\right)\left( 1 - c_{\theta}\right)$\;
	
	\textbf{if} 
			$\textbf{f}_r^{\top}\textbf{b}_{3d}\geq\sqrt{{\left\lVert \textbf{f}_r \right\lVert}^2 - r_{xy}^2(\alpha)}$
	\textbf{then} ~
            $\theta \gets \theta - \frac{\theta_{max}}{2}\frac{1}{2^i}$\;            
   	\textbf{else} ~
            $\theta \gets \theta + \frac{\theta_{max}}{2}\frac{1}{2^i}$\; 
}
\Return $\theta$
}
\end{algorithm}

\subsection{Attitude control}\label{subsec:ctrl:ori}
The attitude controller takes as input the desired orientation computed from the position controller ($\textbf{R}_d$), the measured orientation ($\textbf{R}_B$), and the measured angular speed ($\boldsymbol{\omega}_B$) to compute the desired control torque ($\textbf{u}_{\tau}$).

\subsubsection{Control equations}\label{subsubsec:ctrl:ori:equ}
The desired control torque is computed as 
\begin{equation}\label{eq:ctrlTorque}
\textbf{u}_{\tau} = \boldsymbol{\omega}_B \times \textbf{J} \boldsymbol{\omega}_B - \textbf{K}_R \textbf{e}_R - \textbf{K}_{Ri} \textbf{e}_{Ri} - \textbf{K}_{\omega} \boldsymbol{\omega}_B,
\end{equation}
where $\textbf{K}_R$, $\textbf{K}_{Ri}$, $\textbf{K}_{\omega} \in \mathbb{R}^{3\times3}$ are positive diagonal matrices and $\textbf{e}_R$ is the orientation tracking error defined as 
\begin{equation}
\textbf{e}_R = \frac{1}{2}\left( \textbf{R}_d^{\top} \textbf{R}_B - \textbf{R}_B^{\top} \textbf{R}_d\right)^{\vee},
\end{equation}
with $\bullet^{\vee}$ which is the vee map from $\mathrm{SO}(3)$ to $\mathbb{R}^3$ and $\textbf{e}_{Ri}$ the integral orientation tracking error computed as
\begin{equation}
\textbf{e}_{Ri} = \int_{0}^{t} \textbf{e}_R d\tau.
\end{equation}

\subsection{Wrench mapper}\label{subsec:ctrl:wrc}
The wrench mapper takes as input the desired control force in (\ref{eq:ctrlForce}) and moment in (\ref{eq:ctrlTorque}) provided by the position and attitude controller respectively and computes a feasible $\textbf{u}$ through the nonlinear map
\begin{align}
\textbf{u} &= \textbf{F}(\alpha)^{\ddagger} \begin{bmatrix} \textbf{u}_f \\ \textbf{u}_{\tau} \end{bmatrix},
%w_i^o &= \sqrt{\frac{f_i}{c_f}},
\end{align}
where $\textbf{F}(\alpha) \in \mathbb{R}^{6\times6}$ is the allocation map. Since the structural properties of the allocation map $\textbf{F}(\alpha)$ change with the tilting angle $\alpha$ (\ie with $\alpha = 0$ the allocation map becomes singular or it may be ill-conditioned if $\alpha \approx 0$) the computation of the wrench mapper is not trivial and the use of a simple inversion is not possible.
In \cite{Sima2006} different approaches aimed at modifying the original ill-posed estimation problem with the goal of stabilizing the solution and/or obtaining a meaningful solution are presented, these approaches are known as \textit{regularisation}. 
In particular, the adopted method, known as \textit{Tikhonov regularisation}, computes the solution in closed form as
\begin{equation}
\textbf{F}(\alpha)^{\ddagger} = \left( \textbf{F}(\alpha)^{\top} \textbf{F}(\alpha) + \gamma I_6 \right)^{-1}\textbf{F}(\alpha)^{\top},
\end{equation}
where $\gamma \in\mathbb{R}_{>0}$ is a properly chosen regularization parameter.
Of course, for $\alpha \gg 0$ the allocation matrix is full-rank, then the Tikhonov regularisation is not needed anymore.
It is then mandatory to parametrise $\gamma = \gamma(\alpha)$ in order to make its contribution significant for $\alpha \approx 0$ and negligible for $\alpha \gg 0$.
For this purpose, a hyperbolic curve has been adopted
\begin{equation}
\gamma(\alpha) = \frac{k_1}{\alpha + k_2},
\end{equation}
with $k_1\in\mathbb{R}_{>0}$ and $k_2\in\mathbb{R}_{>0}$ properly chosen.

\subsubsection{Lateral force saturation}\label{subsec:ctrl:wrc:force_sat}
The lateral force achievable by the FAST-Hex increases nonlinearly with the tilting angle $\alpha$.
Let us express the lateral force bound used in the position controller $r_{xy}$ as a function of $\alpha$.
\begin{figure}
\centering
\includegraphics[width=\figWidth\columnwidth]{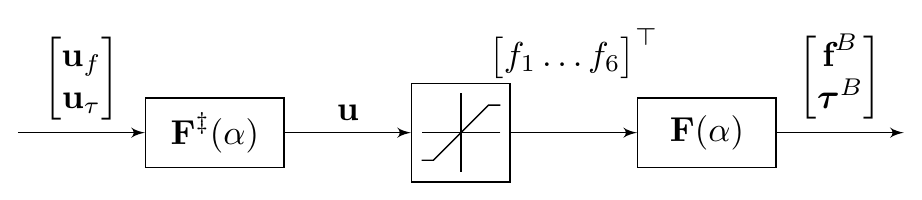}
\vspace{-8mm}
\caption{Wrench allocation with actuators' saturation.} 
\label{fig:force_sat}
\end{figure}

\begin{figure}[htbp]
\centering
\includegraphics[width=\figWidth\columnwidth]{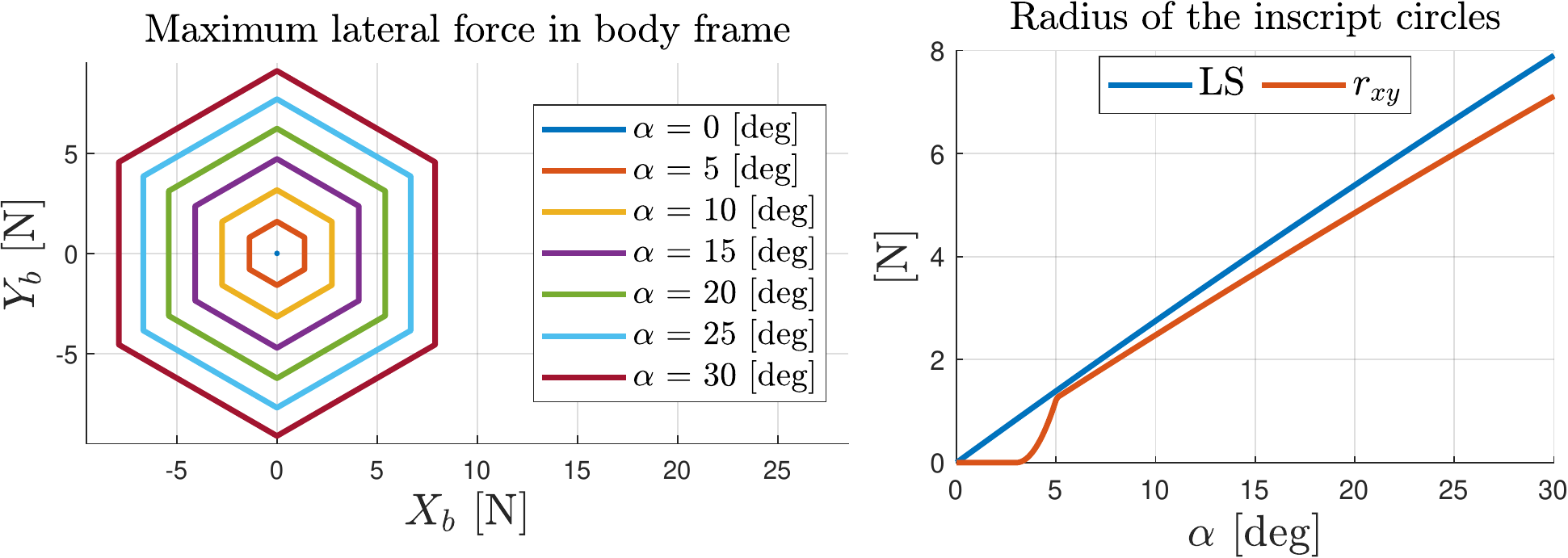}
\vspace{-6mm}
\caption{Left: Saturated lateral force for different values of $\alpha$. Right: Lateral force saturation function.} 
\label{fig:fast_hex_saturation}
\end{figure}

To do so, the scheme represented in Fig.~\ref{fig:force_sat}, in which the propellers' saturation has been taken into account, has been considered.
The maximum achievable lateral force considering as input of the wrench mapper a desired lateral force (\eg $\SI{10}{\newton}$) around the hovering conditions with a null desired moment for different values of $\alpha$ has been then numerically computed.
The obtained saturated force is reported in Fig.~\ref{fig:fast_hex_saturation}-left.
%\begin{figure}[htbp]
%\centering
%\includegraphics[width=0.75\columnwidth]{fast_hex_hexagon.pdf}
%\caption{Saturated lateral force for different values of $\alpha$.} 
%\label{fig:fast_hex_hexagon}
%\end{figure}

Since the sets for admissible planar lateral forces have a hexagonal shape, for the sake of simplicity it has been decided to consider as a lateral bound the circle inscribed in each hexagon. To exploit $r_{xy}$ as a function of $\alpha$ a Least Squares (LS) approach has been used to interpolate the obtained values with a second degree polynomial. 
%The radius of the circles with reference to the tilt angle are reported in Figure \ref{fig:fast_hex_LS}.
Finally, the obtained polynomial has been scaled down with a tunable gain leading to a more conservative lateral force bound.
To cope with the numerical problem related to the ill-conditioned pseudo inverse a dead-zone in the proximity of $\alpha = 0$ has been introduced (see Figure \ref{fig:fast_hex_saturation}-right).
%\begin{figure}[t]
%\centering
%\includegraphics[width=\figWidth\columnwidth]{fast_hex_LS.pdf}
%\caption{Interpolated values of the inscribed circles' radius.} 
%\label{fig:fast_hex_LS}
%\end{figure}

%%%%%%%%%%%%%%%%%%%%%%%%%%%%%%%%%%%%%%%%%%%%%%%%%%%%%%%%%%%%%%%%%%%%%%
\section{Experimental validation}\label{sec:exp}
%%%%%%%%%%%%%%%%%%%%%%%%%%%%%%%%%%%%%%%%%%%%%%%%%%%%%%%%%%%%%%%%%%%%%%

\subsection{Experimental setup}\label{subsec:exp:stp}
The physical and mechanical parameters and controller gains of the FAST-Hex are reported in \tabref{tab:mechParams}.
In particular, the controller gains have been initially tuned on MATLAB/Simulink environment by means of a ad-hoc simulator and eventually fine-tuned on the real flying platform.

%\begin{table}[b]
%	\renewcommand\arraystretch{1.3}
%	\caption{Controller parameters used in the experiments.}
%	\label{tab:FASTHex:ctrl_params}
%	\centering	
%	\resizebox{0.75\columnwidth}{!}{%
%	\begin{tabular}{|CCC|}
%		\hline
%		\text{Parameter} & \text{Value} & \text{Unit} \\
%		\hline
%		\Km_p(j,j)|_{j=1,2,3} & 50, 50, 50 & [\ ] \\
%		\Km_{p_i}(j,j)|_{j=1,2,3} & 20, 20, 20 & [\ ] \\
%		\Km_v(j,j)|_{j=1,2,3} & 14.14, 14.14, 14.14 & [\ ] \\
%		\Km_R(j,j)|_{j=1,2,3} & 15, 15, 6 & [\ ] \\
%		\Km_{R_i}(j,j)|_{j=1,2,3} & 1, 1, 1 & [\ ] \\
%		\Km_w(j,j)|_{j=1,2,3} & 1.5, 1.5, 0.5 & [\ ] \\
%		\hline
%		\overline{\ev}_{p_i}=-\underline{\ev}_{p_i} & [3\ 3\ 3]^{\top} & \si{\meter} \\
%		\overline{\ev}_{R_i}=-\underline{\ev}_{R_i} & [1\ 1\ 1]^{\top} & \si{rad} \\
%		\hline
%		n_{\text{it}} & 20 & [\ ] \\
%		\red{\overline{w_i}} & \red{98} & \red{\si{Hz}} \\
%		\red{\underline{w_i}} & \red{20} & \red{\si{Hz}} \\
%		\multicolumn{3}{|c|}{\db{Mention that we made conservative choices.}} \\
%		\overline{\dot{\alpha}}=-\underline{\dot{\alpha}} & 10 & \si{\degree\per\second} \\
%		\hline
%	\end{tabular}	
%} %resizebox
%\end{table}

\begin{figure*}[t]
	\centering
	\includegraphics[width=\figWidth\columnwidth]{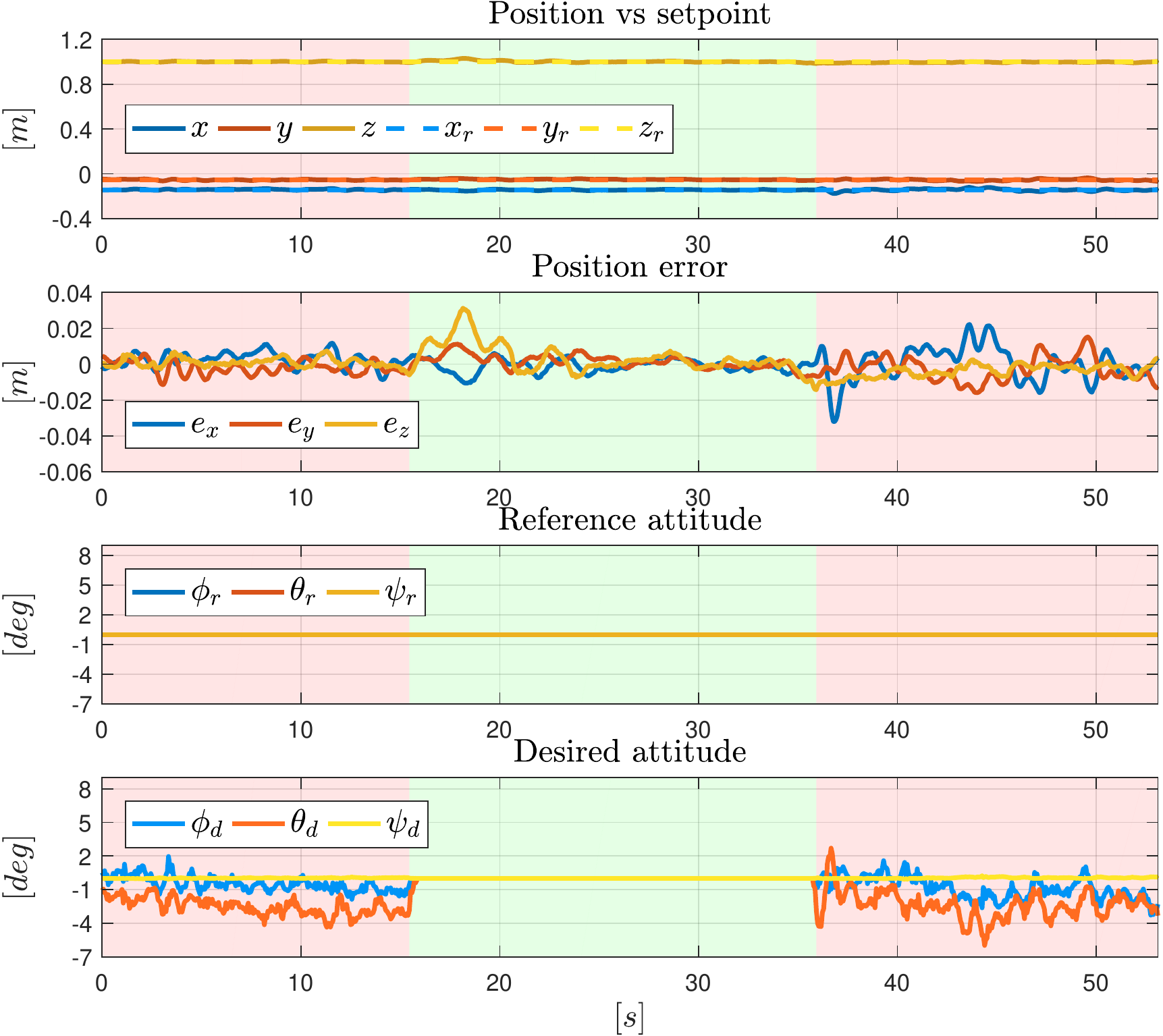}\hfill
	\includegraphics[width=\figWidth\columnwidth, trim=0 0.075cm 0 0cm]{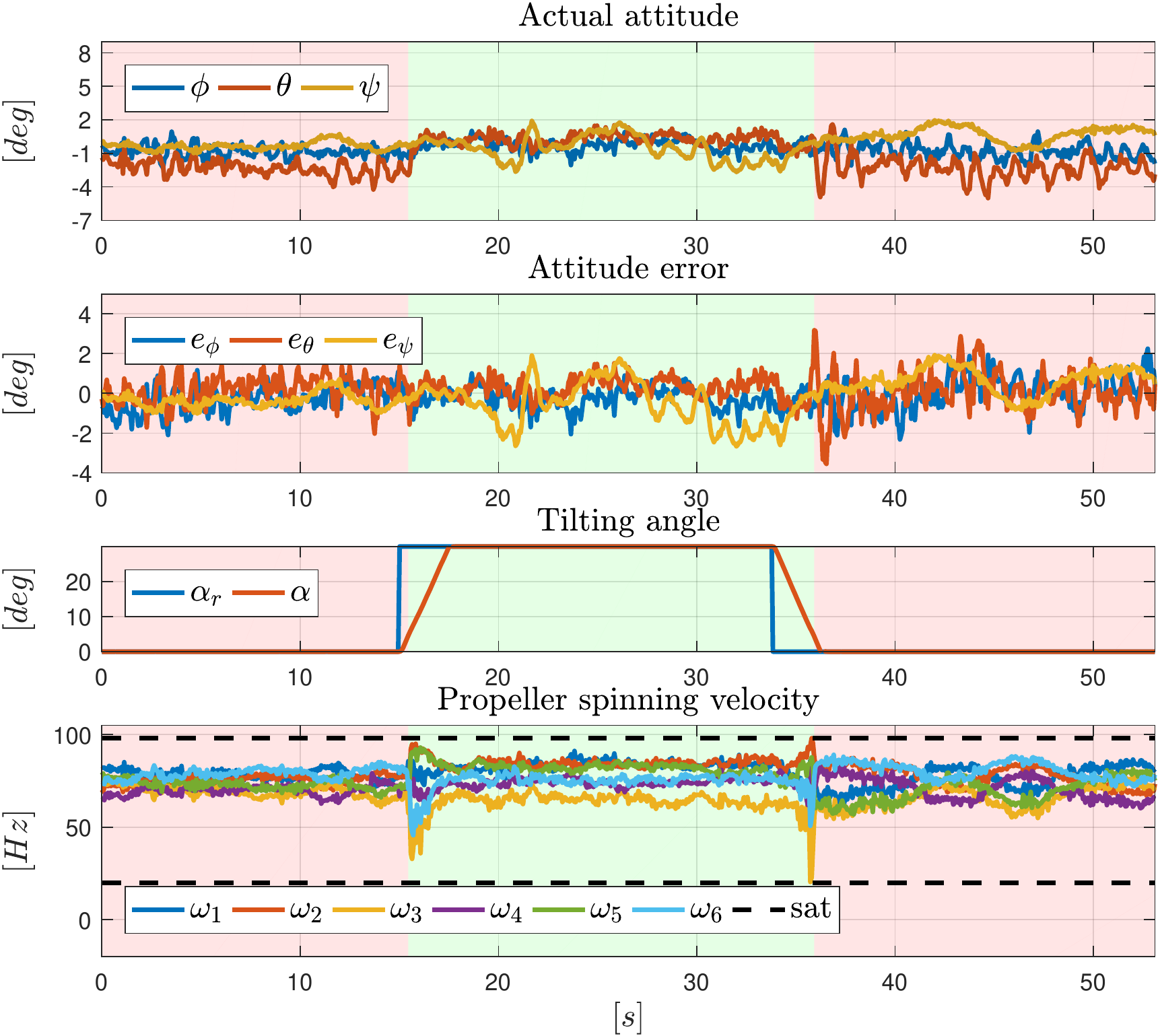}
	\vspace{-2mm}
	\caption{Plots of Experiment 1 - left column from top to bottom. 
	1) Actual vs reference position; 
	2) Position tracking error; 
	3) Reference attitude depicted in Euler angles; 
	4) Desired attitude depicted in Euler angles.
	Right column from top to bottom: 
	5) Actual attitude depicted in Euler angles; 
	6) Attitude tracking error; 
	7) Reference and actual tilting angle; 
	8) Actual propeller spinning velocity.
	While the FAST-Hex is under-actuated the plots are highlighted in red. 
	On the other hand, during full actuation the plots are highlighted in green.} 
	\label{fig:alpha_step_1:AllPlots}
	\vspace{-3mm}
\end{figure*}

The controller has been developed in Matlab-Simulink and runs at a frequency of $\SI{500}{\hertz}$ on a stationary ground station. 
The ground station is connected with the FAST-Hex with a serial cable. 
This setup has been selected for fast development and testing of the controller but could be ported with some straightforward effort to an on-board system as the computational demand of the controller is negligible. 
Therefore we would expect an increased performance as an on-board control would benefit from a possibly higher control frequency, no communication delay and no disturbance from the hanging serial cable. 
The following presented experiments are therefore a baseline on which the system could be improved.

On-board the FAST-Hex an inertial measurement unit provides acceleration and angular rate at $\SI{500}{\hertz}$. 
An external marker-based motion capture (MoCap) system provides with sub-centimeter accuracy the pose measurements of the aerial robot at $\SI{100}{\hertz}$. 
The IMU and the pose measurements are fused via an Unscented Kalman Filter state estimator to obtain full state estimates at control frequency rate ($\SI{500}{\hertz}$). 
The external MoCap system could as well be replaced by an on-board camera and a Perspective-n-Point algorithm to estimate the robot's pose.
However, we purposefully neglected this possibility to evaluate the FAST-Hex and its controller without additional influences of the particular perception system. 

We report two sets of experiments in this paper. 
In the first set (see \secref{subsec:exp:exp1}) we demonstrate basic hovering capabilities during reconfiguration of the tilting angle. 
In the second set (see \secref{subsec:exp:exp2}) we present dynamic trajectory tracking for two kinds of trajectories, sinusoidal attitude tracking with a fixed position and sinusoidal position tracking with a fixed attitude, both with a time varying tilting angle. 
{An additional experiment, comparing the robustness of the platform to external force disturbance during full- and under-actuation, can be found in the attached technical report.}
We will present several plots in the following figures. 
In single column figures we refer to the plots from top to bottom with increasing numbers. 
In double column figures we refer to the plots from top to bottom in the first column and then from top to bottom in the second column with increasing numbers. 
For an easier understanding, we highlighted in all plots with a bright red background while the FAST-Hex is in UDT-configuration and with a bright green or yellow background as soon as the platform is in MDT-configuration.
In order to better appreciate the discussed experiments and their results, we suggest the reader to watch the attached videos.

%	write here
% frequency, POM, battery, sensors used (what used and why), two filters for the state estimation, calibration of the two parameters in the allocation matrix, \etc

\subsection{Experiment 1: Static Hovering}\label{subsec:exp:exp1}
In this experiment, the FAST-Hex is commanded to hover statically, \ie to resist the gravitational force while maintaining a constant position $\pv_r = [-0.14\ -0.05\ 1]^{\top}\SI{}{\meter}$ and a horizontal orientation, \ie $\Rm_r=\eye{3}$. 
Additionally, the reference angle $\alpha_r$ for the synchronized tilting angle of the actuators has a rectangular profile between the values $\alpha_1 = \ang{0}$ and $\alpha_2 = \ang{30}$ (see first, third and seventh plot in \figref{fig:alpha_step_1:AllPlots}). 
As a consequence, the robot switches its configuration from \gls{udt} to \gls{mdt} and back.

The goal of the experiment is to demonstrate the controller's capability to safely change between the two configurations UDT and MDT and assess the controller's robustness with respect to the unmodeled effects discussed in \secref{sec:model_discussion}.

Observing the position and attitude error plots (plot 2 and 5 in \figref{fig:alpha_step_1:AllPlots}) it is obvious that the controller copes very well with the configuration transition. 
Generally, the overall mean position tracking error is $\overline{\lVert\vect{e}_p\rVert} = \SI{8.7}{\milli\meter}$, with a significantly smaller tracking error while being in \gls{mdt} configuration ($\overline{\lVert\vect{e}_p^\text{\gls{mdt}}\rVert} = \SI{5.5}{\milli\meter}$ vs. $\overline{\lVert\vect{e}_p^\text{\gls{udt}}\rVert} = \SI{6.7}{\milli\meter}$ - we ignored the initial time after a configuration change as the transition causes a short increase of the tracking error). 
The overall mean attitude tracking error is as well small ($\overline{e_\phi} = \SI{0.84}{\degree}$, $\overline{e_\theta} = \SI{0.92}{\degree}$, $\overline{e_\psi} = \SI{1.10}{\degree}$) with again a significantly smaller mean error for roll and pitch during \gls{mdt} configuration ($\overline{e_\phi^\text{\gls{mdt}}} = \SI{0.35}{\degree}$, $\overline{e_\theta^\text{\gls{mdt}}} = \SI{0.35}{\degree}$ vs. $\overline{e_\phi^\text{\gls{udt}}} = \SI{0.54}{\degree}$, $\overline{e_\theta^\text{\gls{udt}}} = \SI{0.45}{\degree}$). 
However, the yaw tracking error is larger ($\overline{e_\psi^\text{\gls{mdt}}} = \SI{0.76}{\degree}$ vs. $\overline{e_\psi^\text{\gls{udt}}} = \SI{0.37}{\degree}$). 
This is due to small misalignments of the propellers, whose effects on the tracking performance are more evident when the control authority on the yaw moment is larger, \ie when $\alpha >>0$.

The plots of the reference $\vect{R}_r$ and the desired attitude $\vect{R}_d$ (see \figref{fig:alpha_step_1:AllPlots} - plot 3 and 4) induce interesting insights on the behavior of the inner attitude control loop. 
Comparing the third and the fourth plot, it becomes clear that the control algorithm is required to re-compute the desired orientation for the system while being in UDT configuration. 
Indeed, when the system is under-actuated the only feasible reference is the one given by the well-known flatness property~\cite{2018c-FaeFraSca}. 
In this case, the desired orientation is continuously regulated to correct position errors. 
The non-exact zero mean for $\phi_d$ and $\theta_d$ is due to parameters mismatches between the model and the real system, especially of those associated with the orientation of the actuators, and to external disturbances like the one induced by the serial cable. 
%This is confirmed by plot seventh of \figref{fig:alpha_step_1:AllPlots}. 
%Notably, \gls{mrav} applies a small force along the $\xB$ axis with a negative non-zero mean.
Conversely, as soon as the angle $\alpha$ is large enough the robot can exert lateral forces without the need of re-orient itself and so the desired attitude can be constantly flat. 

Finally we would like to discuss the desired spinning velocities for the rotors computed by the pose controller depicted in the sixth plot. 
As it can be appreciated, the signals remain bounded by their limits, which demonstrate the controller ability to comply with the actuator bounds. 
Furthermore, it is worthwhile to observe the peaks in the actuator commands during the changes of configuration, due to the crossing of the singularity discussed in Sec~\ref{subsec:ctrl:wrc} that is also the cause of the increase in the position tracking error.

\subsection{Experiment 2: Dynamic Trajectory Tracking}\label{subsec:exp:exp2}
In this set of experiments, we command the FAST-Hex to track two trajectories with independent position and orientation profile, which is clearly unfeasible for standard collinear multirotor platforms. 
As in the previous experiment, we altered the tilting angle over time. 
The goal of these two experimental sets is to demonstrate how the pose tracking of the controller is fulfilled when the tilting angle is changed over time.

\subsubsection{Sinusoidal translation with constant horizontal attitude}
\begin{figure}[t]
	\centering
	\figExp2_Image
	\caption{Time-lapse pictures of the FAST-Hex during Experiment 2-a: Top: While being in UDT-configuration (tilting angle is zero) the platform cannot generate horizontal forces and the controller needs to adapt the attitude trajectory to be able to track the position trajectory. 
	Bottom: With tilted propellers, the FAST-Hex is able to generate lateral forces and the platform can track independent position and attitude trajectories (depending on the actuation constraints) and therefore remain horizontal while traversing laterally.} 
	\label{fig:Exp2Image}
\end{figure}

\begin{figure}[th]
	\centering
	\includegraphics[width=\figWidth\columnwidth]{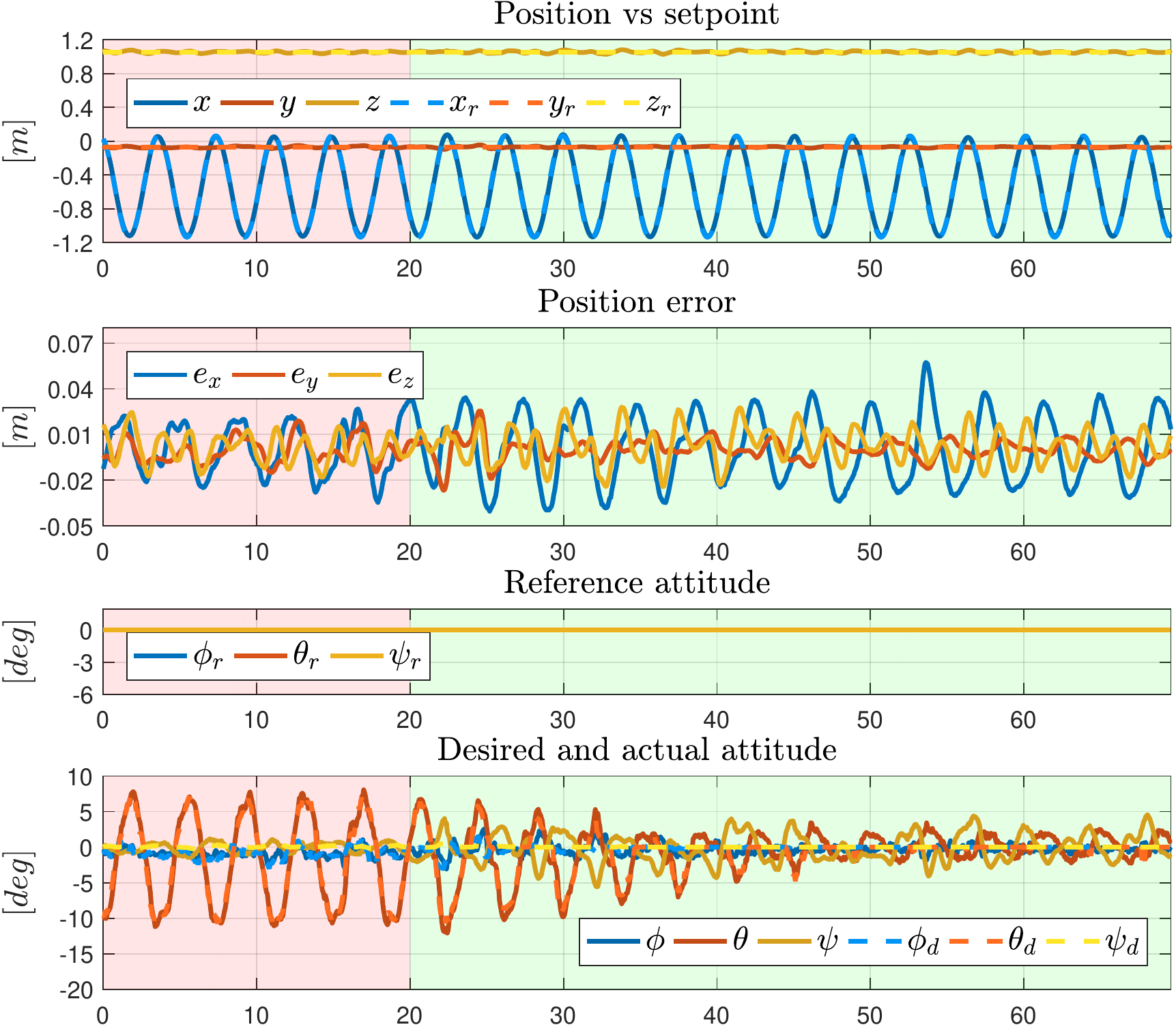}\\
	\includegraphics[width=\figWidth\columnwidth]{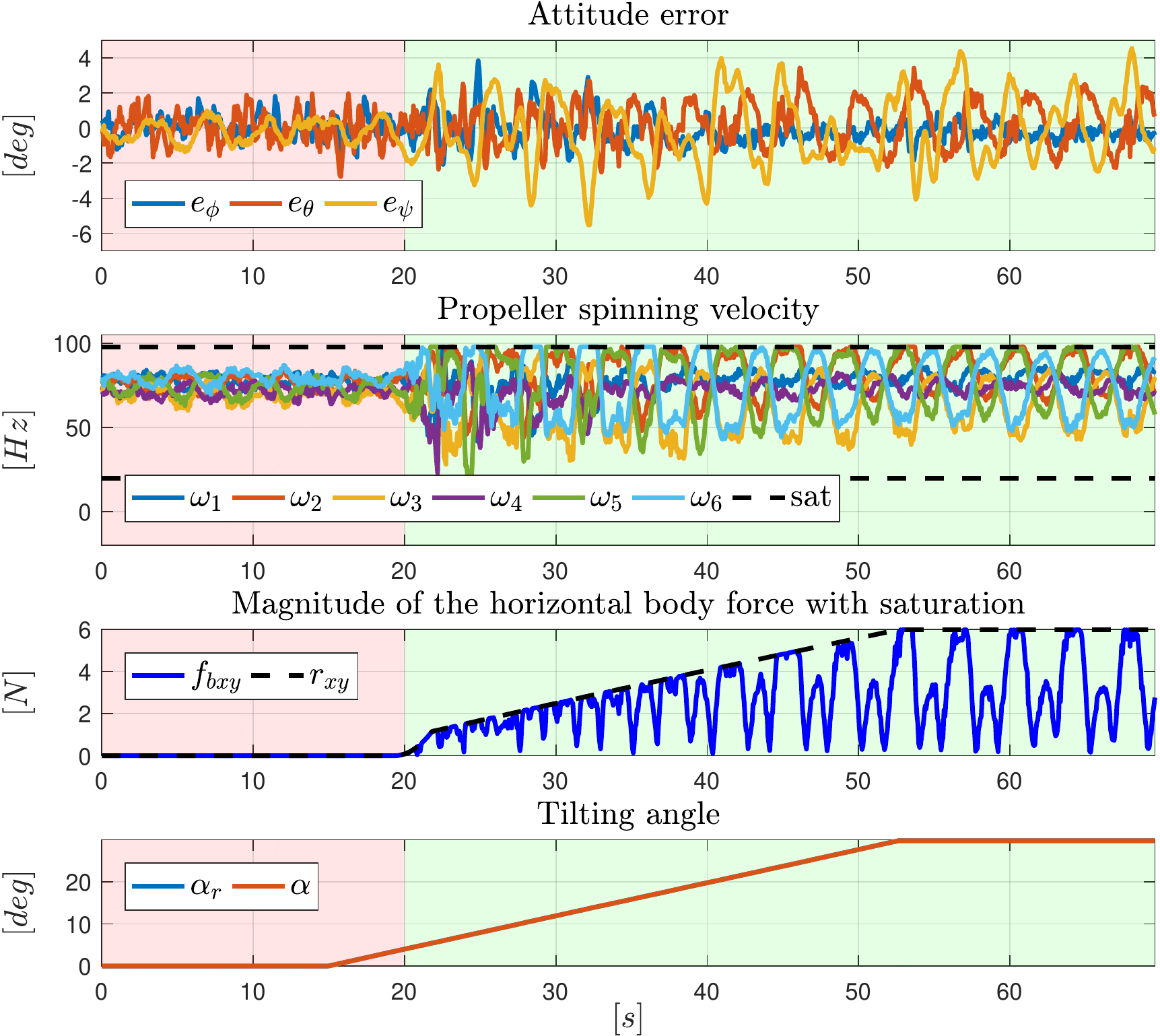}
	\vspace{-7mm}
	\caption{Plots of Experiment 2-a - from top to bottom. 
	1) Actual vs reference position; 
	2) Position tracking error; 
	3) Reference attitude depicted in Euler angles; 
	4) Desired and actual attitude depicted in Euler angles;
	5) Attitude tracking error; 
	6) Actual propeller spinning velocity; 
	7) Maximum and actual lateral force; 
	8) Reference and actual tilting angle.
	While the FAST-Hex is under-actuated, the plots are highlighted in red, during full actuation, the plots are highlighted in green.} 
	\label{fig:pos_sin_3:stateset}
\end{figure}

In this experiment, we aim at tracking a translational sine-wave trajectory while maintaining a horizontal attitude ($\Rm_r=\eye{3}$). 
The amplitude of the translational sine-wave is $\SI{1.2}{\meter}$ with a peak velocity of $\hat{\dot p}_{x_r} = \SI{1}{\meter\per\second}$ and a peak acceleration of $\hat{\ddot p}_{x_r} = \SI{1.67}{\meter\per\square\second}$ (see \figref{fig:pos_sin_3:stateset} - 1). 
The tilting angle $\alpha$ is increased over time from $\SI{0}{\degree}$ to $\SI{30}{\degree}$  (see \figref{fig:pos_sin_3:stateset} - last plot). 
A photograph of the FAST-Hex while tracking this trajectory in the two different configurations is provided in \figref{fig:Exp2Image}.

From plots 3 and 4 of \figref{fig:pos_sin_3:stateset} it becomes clear that the controller has to significantly alter the reference trajectory to output a trackable desired trajectory while the platform is under-actuated (until $t\approx \SI{20}{\second}$). 
During this initial phase of the experiment, the maximum lateral force $f_{xy}$ is zero (see plot 7) making the attitude trajectory fully coupled with the position trajectory. 
As soon as the lateral force $f_{xy}$ is not zero but increases over time, the desired trajectory gradually approaches the reference trajectory. 
It is interesting to point out that even with fully tilted propellers, the lateral forces required to track a fully horizontal trajectory would violate the maximum spinning velocity of the propellers (see plot 6). 
Therefore, the desired trajectory diverges slightly from the reference trajectory at the peaks of the translation.

\subsubsection{Hovering with sinusoidal rolling}
\begin{figure}[t]
	\centering
	\includegraphics[width=\figWidth\columnwidth]{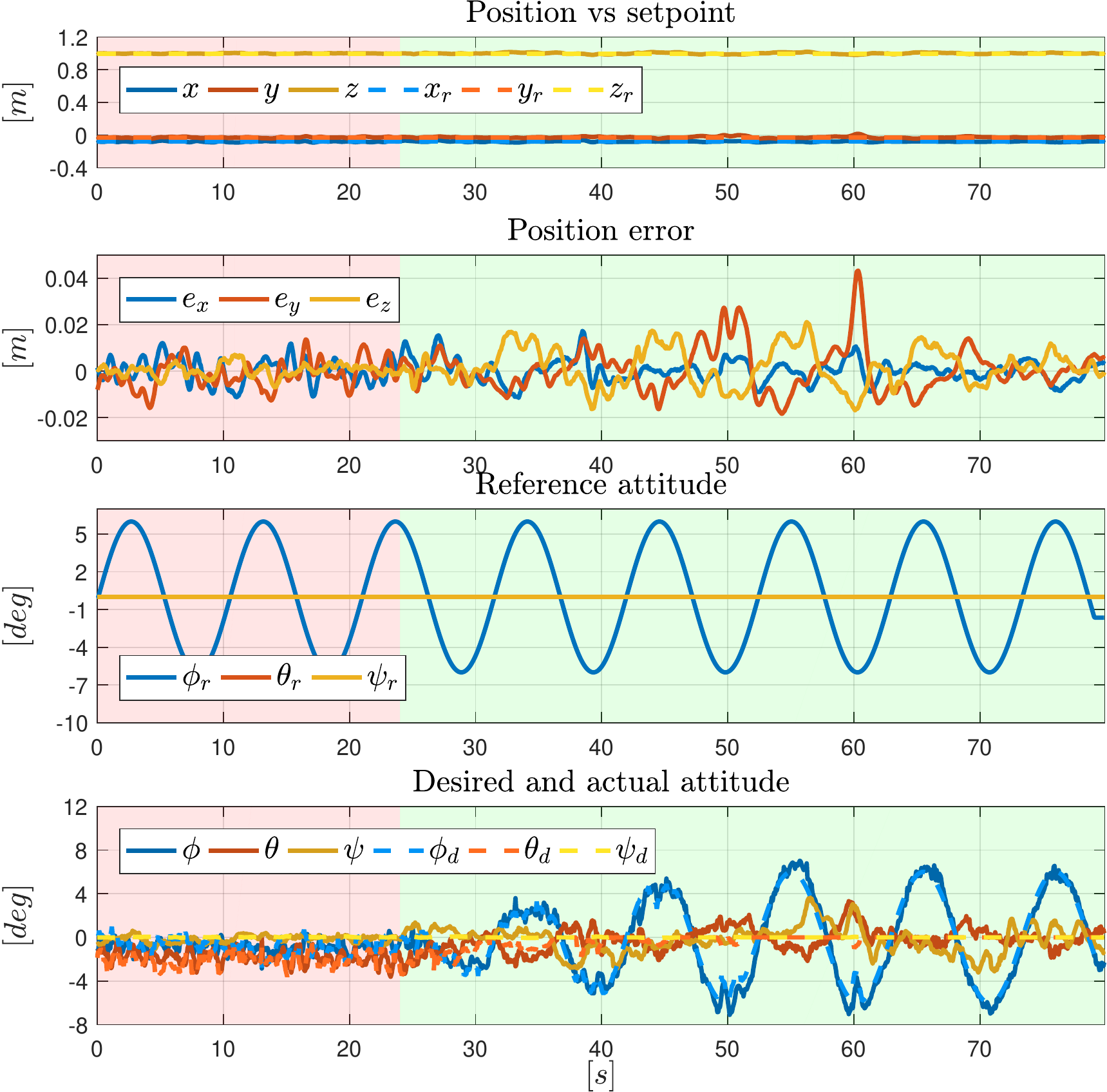} \vspace{-6.5mm}\\%\hfill
	\includegraphics[width=\figWidth\columnwidth, trim=0 -0.5cm 0 0cm]{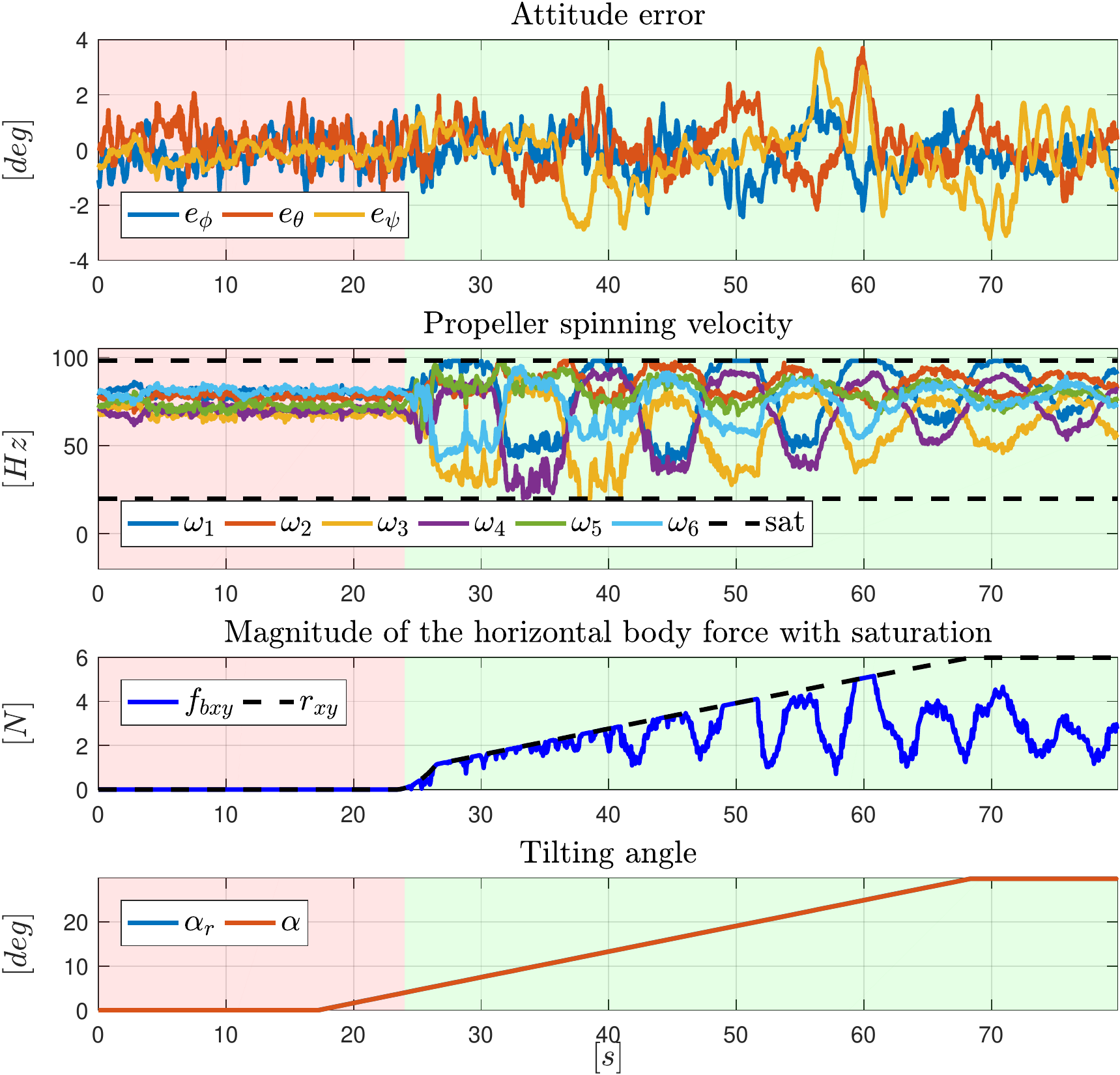}
	\vspace{-10mm}
	\caption{Plots of Experiment 2-b - from top to bottom. 
	1) Actual vs reference position; 
	2) Position tracking error; 
	3) Reference attitude depicted in Euler angles; 
	4) Desired and actual attitude depicted in Euler angles;
	5) Attitude tracking error; 
	6) Actual propeller spinning velocity; 
	7) Maximum and actual lateral force; 
	8) Reference and actual tilting angle.
	While the FAST-Hex is under-actuated, the plots are highlighted in red, during full actuation, the plots are highlighted in green.} 
	\label{fig:roll_sin_1:stateset}
\end{figure}

In the second dynamic reference motion, the position trajectory is constant with $\vect{p}_r = [-0.08\ -0.03\ 1]^{\top}\SI{}{\meter}$, while the roll angle follows a sine-wave with a peak angle of $\SI{6}{\degree}$ and a frequency of about $\SI{0.1}{\hertz}$. The pitch and yaw angles remain constant at $\SI{0}{\degree}$. 
The plots related to this trajectory, which is clearly unfeasible for a UDT vehicle, are depicted in \figref{fig:roll_sin_1:stateset}. 

%The reference tilting angle increases linearly from $\alpha=\SI{0}{\degree}$ at $t = \SI{17.25}{\second}$ until it reaches $\alpha=\SI{30}{\degree}$ at $t=\SI{29.75}{\second}$. 
%Then, it maintains this value until the end of the experiment (see last plot in \figref{fig:roll_sin_1:stateset}). 
The reference tilting angle is increased linearly from $\alpha=\SI{0}{\degree}$ to $\alpha=\SI{30}{\degree}$, as in the previous experiment (see last plot in \figref{fig:roll_sin_1:stateset}). 
The controller is therefore required to adapt the reference trajectory into a trackable desired trajectory.

The static reference $\vect{p}_r$ and actual body position $\vect{p}_B$ are depicted in the first plot in \figref{fig:roll_sin_1:stateset}.
The position tracking error remains small with mean position error of $\overline{\lVert\vect{e}_p\rVert} = \SI{10.4}{\milli\meter}$. 
The position error does not significantly change between the configurations. 
A standard collinear multirotor is not able to track a trajectory for roll and pitch while remaining at a fixed location, as multi-directional forces would need to be applied. 
Therefore, the FAST-Hex cannot track the reference attitude trajectory initially (see \figref{fig:roll_sin_1:stateset} - plot 4). 
As a consequence, the controller outputs a desired trajectory that is basically constant and horizontal. 
As soon as the feasible horizontal body force is large enough (see plot 7) thanks to an increasing tilting angle, the FAST-Hex gradually starts to track the reference attitude trajectory. 
Starting from $t\approx\SI{60}{\second}$, the tilting angle is large enough to fully track the reference attitude. 
The lateral forces required to track the desired rolling motion can now be completely generated by the propellers (see plot 6 and 7).

\section{Conclusion and Future Work}\label{sec:conclusion}
%%%%%%%%%%%%%%%%%%%%%%%%%%%%%%%%%%%%%%%%%%%%%%%%%%%%%%%%%%%%%%%%%%%%%%
In this paper, we presented a novel morphing hexarotor platform - the FAST-Hex. 
The careful integration of a single additional actuator allows the platform to efficiently transition from under-actuation to full-actuation. 
We presented and discussed the hardware implementation, and the control framework that allows to drive the platform seamlessly in both conditions, while prioritizing position tracking over attitude tracking if the actuation limitations cannot be met otherwise.

We presented an extensive set of flight experiments, showing general trajectory tracking performance in static and dynamic flight regimes in both configurations. 
Furthermore, we discussed the benefits of morphing aerial platforms under the effect of external force disturbances.

In the future we plan to compare the use of alternative controllers based on online optimization to the current proposed solution.

%%%%%%%%%%%%%%%%%%%%%%%%%%%%%%%%%%%%%%%%%%%%%%%%%%%%%%%%%%%%%%%%%%%%%%%
%\section*{ACKNOWLEDGEMENTS}\label{sec:ack}
%%%%%%%%%%%%%%%%%%%%%%%%%%%%%%%%%%%%%%%%%%%%%%%%%%%%%%%%%%%%%%%%%%%%%%
%We thank Anthony Mallet (LAAS-CNRS) for his contribution in the software architecture of the experiments.

%%%%%%%%%%%%%%%%%%%%%%%%%%%%%%%%%%%%%%%%%%%%%%%%%%%%%%%%%%%%%%%%%%%%%%
\bibliographystyle{IEEEtran}
% DO NOT ERASE THE NEXT LINE,
% ONLY COMMENT IT AND DECOMMENT THE NEXT-NEXT, IF YOU NEED
\bibliography{bibAlias,bibMain,bibNew,bibAF,./bibCustom}
%%%\bibliography{./bibCustom}
%%%%%%%%%%%%%%%%%%%%%%%%%%%%%%%%%%%%%%%%%%%%%%%%%%%%%%%%%%%%%%%%%%%%%%
\vspace{-12mm}
%% Markus
\begin{IEEEbiography}{Markus Ryll}
obtained a Diploma in Mechatronics in 2008 and a Master Degree in medical engineering in 2010. He received the Ph.D. degree from the Max Planck Institute for Biological Cybernetics in T\"ubingen, Germany in cooperation with the University of Stuttgart, Germany in 2015. From 2014 to 2017 Markus was a Research Scientist at the RIS team at LAAS-CNRS, Toulouse, France. Since 2018 Markus is a Senior Research Scientist at the Robust Robotics Group at the Massachusetts Institute of Technology, Cambridge, USA. %Markus Ryll's main research interests are design modeling and control of novel robotic systems, mainly focused on aerial platforms. 
\end{IEEEbiography}
\vspace{-12mm}

%% Davide
\begin{IEEEbiography}{Davide Bicego} is a Post-Doctoral Researcher at the University of Twente, Enschede, The Netherlands, in the group of Robotics and Mechatronics (RAM). From 2016 to 2019, he carried out a Ph.D. at the Laboratoire d'Analyse et d'Architecture des Systèmes (LAAS-CNRS), Toulouse, France, in the Robotics and Interactions (RIS) group. He received the B.Sc. and the M.Sc. degrees in Information Engineering and Automation Engineering in 2013 and 2015, respectively, from University of Padua, Padua, Italy.
%%%   IF THERE IS ENOUGH SPACE YOU CAN DE-COMMENT THE FOLLOWING LINE   %%%
%Davide does research in Aerial Robotics \& Physical Interactions exploiting Multi-Directional Thrust Aerial Vehicles (MDT-UAVs), Control \& Mechanical Engineering and Computer Engineering. 
\end{IEEEbiography}
\vspace{-12mm}

%% Mattia
\begin{IEEEbiography}{Mattia Giurato} received the B.Sc. and the M.Sc. degrees in Automation and Control Engineering (in 2013 and 2015 respectively) from Politecnico di Milano and he concluded in 2020 his Ph.D. in Aerospace Engineering in the Aerospace Science and Technology department of Politecnico di Milano. He is now a Post-Doctoral Researcher in the Aerospace System and Control Laboratory (ASCL).%His research interests include: model identification, robust control, nonlinear control, adaptive control, UAV design.
\end{IEEEbiography}
\vspace{-12mm}

%% Marco
\begin{IEEEbiography}{Marco Lovera}
(M98) is a Professor of Automatic Controls at the Politecnico di Milano. After a one-year period in industry he joined in 1999 the Dipartimento di Elettronica, Informazione e Bioingegneria of the Politecnico di Milano. Since 2015 he is with the Dipartimento di Scienze e Tecnologie Aerospaziali of the Politecnico di Milano, where he leads the Aerospace Systems and Control Laboratory (ASCL). %Main research interests: model identification and aerospace applications of advanced systems and control concepts, with specific reference to spacecraft attitude determination and control, rotorcraft modelling and control, small scale autonomous vehicles.
\end{IEEEbiography}
\vspace{-12mm}

%% Antonio
\begin{IEEEbiography}{Antonio Franchi} (S’07-M’11-SM’16)
is a Professor of Robotics in the Faculty of Electrical Engineering, Mathematics \& Computer Science, at the University of Twente, Enschede, The Netherlands, and an Associate Researcher at LAAS-CNRS, Toulouse, France. His main research interests include the design and control for robotic systems with applications to multi-robot systems and aerial robots. He co-authored more than 130 papers in peer-reviewed international journals and conferences. He is an IEEE Senior Member. %Since 2012 he has been serving in the editorial board of the major IEEE robotics conferences and journals, including the IEEE Transactions on Robotics. He is the co-founder and co-chair of the IEEE RAS Technical Committee on Multiple Robot Systems.
\end{IEEEbiography}
\end{document}

%% file: macros/boldfacemath.tex
\newcommand{\vect}[1]{\mathbf{#1}}
\newcommand{\mat}[1]{\mathbf{#1}}

\newcommand{\diffs}[3]{\frac{\partial^2 #1}{
\ifx#2#3 
\partial #2^2
\else
\partial #2 \partial #3
\fi
}}

\newcommand{\bs}{\boldsymbol}
\newcommand{\rank}{\mathrm{rank}}

\newcommand{\eye}[1]{\mathbf{I}_{#1}}

\newcommand{\pv}{\vect{p}}

\newcommand{\Jm}{\mat{J}}

\newcommand{\Km}{\mat{K}}

\newcommand{\Rm}{\mat{R}}

\newcommand{\wFrame}{\mathcal{F}_W}		% word frame
\newcommand{\bFrame}{\mathcal{F}_B}		% body frame

%% file: macros/glossaries.tex
\newacronym[longplural={Multi-Rotor Aerial Vehicles}]{mrav}{MRAV}{Multi-Rotor Aerial Vehicle}			

\newacronym{aphi}{APhI}{Aerial Physical Interaction}

\newacronym{udt}{UDT}{Uni-Directional Thrust}
\newacronym{mdt}{MDT}{Multi-Directional Thrust}
\newacronym{od}{OD}{Omni-Directional}
\newacronym{fra}{FRA}{Fully-Ranked Allocation Matrix}
\newacronym{lbf}{LBF}{Laterally-Bounded input Force}

\newacronym[longplural={Degrees of Freedom}]{dof}{DoF}{Degree of Freedom}
\newacronym{com}{CoM}{Center of Mass}
\newacronym[longplural={Electronic Speed Controller}]{esc}{ESC}{Electronic Speed Controller}
\newacronym{bldc}{BLDC}{Brushless DC}
\newacronym{pwm}{PWM}{Pulse Width Modulation}	

\newacronym{imu}{IMU}{Inertial Measurement Unit}
\newacronym{gnss}{GNSS}{Global Navigation Satellite System}
\newacronym{gps}{GPS}{Global Positioning System}
\newacronym{mocap}{MoCap}{Motion Capture System}
\newacronym{cad}{CAD}{Computer Aided Design}

\newacronym{ros}{ROS}{Robot Operating System}

\newacronym{pid}{PID}{Proportional, Integrative and Derivative}

%% file: symbol_definitons.tex
\makeatletter
\renewcommand{\dddot}[1]{%
  {\mathop{\kern\z@#1}\limits^{\vbox to-1.4\ex@{\kern-\tw@\ex@
   \hbox{\normalfont ...}\vss}}}}
\renewcommand{\ddddot}[1]{%
  {\mathop{\kern\z@#1}\limits^{\vbox to-1.4\ex@{\kern-\tw@\ex@
   \hbox{\normalfont....}\vss}}}}
\makeatother

\renewcommand{\vect}[1]{\mathbf{#1}}
\newcommand{\matr}[1]{\mathbf{#1}}

\newcommand{\nR}[1]{\mathbb{R}^{#1}}		% real

\DeclareMathOperator{\atantwo}{atan2}

\newcommand{\vx}{\vect{x}}
\newcommand{\vy}{\vect{y}}

\newcommand{\vz}{\vect{z}}

\newcommand{\rotMatB}{\matr{R}_B}		% rotation matrix
\newcommand{\vomegaB}{\boldsymbol{\omega}_B}	% vector omega